%% file: main.tex
\documentclass[sigconf]{acmart}

\AtBeginDocument{
  \providecommand\BibTeX{{
    Bib\TeX}}}
    
\usepackage{amsmath,amsfonts}
\usepackage{algorithmic}
\usepackage{graphicx}
\usepackage{textcomp}
\usepackage{xcolor}
\def\BibTeX{{\rm B\kern-.05em{\sc i\kern-.025em b}\kern-.08em
    T\kern-.1667em\lower.7ex\hbox{E}\kern-.125emX}}
    
\usepackage{hyperref}
\newcommand{\codeurl}[1]{\leavevmode\href{#1}{\textrm{\nolinkurl{#1}}}}
\usepackage{booktabs}
\usepackage{multirow}
\usepackage{wrapfig}
\usepackage{makecell}
\usepackage{algorithm}

\usepackage[switch]{lineno}

\setcopyright{acmlicensed}
\copyrightyear{2018}
\acmYear{2018}
\acmDOI{XXXXXXX.XXXXXXX}
\acmConference[Conference acronym 'XX]{Make sure to enter the correct
  conference title from your rights confirmation email}{June 03--05,
  2018}{Woodstock, NY}
\acmISBN{978-1-4503-XXXX-X/2018/06}

\begin{document}

\title{Preserving Clusters in Error-Bounded Lossy Compression of Scientific Particle Data}

\author{Congrong Ren}
\email{ren.452@osu.edu}
\orcid{0009-0006-6285-7271}
\affiliation{
  \institution{The Ohio State University}
  \city{Columbus}
  \state{Ohio}
  \country{USA}}

\author{Sheng Di}
\email{sdi1@anl.gov}
\orcid{0000-0002-7339-5256}
\affiliation{
  \institution{Argonne National Laboratory}
  \city{Lemont}
  \state{Illinois}
  \country{USA}}

\author{Katrin Heitmann}
\email{heitmann@anl.gov}
\orcid{0000-0003-1468-8232}
\affiliation{
  \institution{Argonne National Laboratory}
  \city{Lemont}
  \state{Illinois}
  \country{USA}}

\author{Franck Cappello}
\email{cappello@mcs.anl.gov}
\orcid{0000-0002-7890-3934}
\affiliation{
  \institution{Argonne National Laboratory}
  \city{Lemont}
  \state{Illinois}
  \country{USA}}

\author{Hanqi Guo}
\email{guo.2154@osu.edu}
\orcid{0000-0001-7776-1834}
\affiliation{
  \institution{The Ohio State University}
  \city{Columbus}
  \state{Ohio}
  \country{USA}}

\renewcommand{\shortauthors}{Ren et al.}

\begin{abstract}
Scientific particle simulations in cosmology, molecular dynamics, and fluid dynamics produce large-scale datasets whose storage, movement, and analysis increasingly rely on lossy compression. However, existing compressors typically bound only pointwise position errors, providing no guarantee on the fidelity of structures derived from particle coordinates, such as single-linkage clustering (also known as Friends-of-Friends algorithm), where clusters are connected components of a proximity graph formed by linking particle pairs within a distance threshold. Even small coordinate perturbations near this threshold can break true links or create false links, thereby splitting or merging entire clusters. We propose a compressor-independent correction technique for preserving single-linkage cluster membership under lossy compression. Our method operates on reconstructed outputs from off-the-shelf compressors such as SZ3, ZFP, Draco, and LCP, and stores a compact corrective edit stream. Our key observation is that cluster-membership queries depend on connected components rather than the complete set of proximity links. Based on this observation, we introduce three constraint-selection modes, vulnerable-pair, safe-component, and halo-forest, that progressively reduce the constraints enforced during correction. Projected gradient descent then corrects the reconstructed coordinates to eliminate the selected violations while respecting the original pointwise error bound. Experiments on cosmology, molecular dynamics, and fluid dynamics datasets with single-GPU and distributed-memory implementations show that our method preserves cluster membership while improving compression ratio by up to 4$\times$ and maintains competitive end-to-end throughput compared to the same base compressors configured with sufficiently tight error bounds to preserve clustering. All code is available at \codeurl{https://github.com/SIGMOD2027Submission/ParticleCluster}.
\end{abstract}


\received{20 February 2007}
\received[revised]{12 March 2009}
\received[accepted]{5 June 2009}

\maketitle

\input{1.intro}
\input{2.related}
\input{3.method}
\input{4.evaluation}
\input{5.conclusion}

\begin{acks}
The material was supported by the U.S. Department of Energy, Office of Science, Advanced Scientific Computing Research (ASCR), under contracts DE-AC02-06CH11357 and DE-SC0025677. This research used resources of the National Energy Research Scientific Computing Center (NERSC), a Department of Energy User Facility using NERSC award DDR-ERCAP 0034457.
\end{acks}

\bibliographystyle{ACM-Reference-Format}
\bibliography{references}

\end{document}

%% file: 1.intro.tex
\section{Introduction}
\label{sec:intro}
Scientific simulations have become one of the largest and fastest-growing sources of data in modern computing, posing acute data management challenges in storage, movement, and analysis. In cosmology, molecular dynamics, and fluid dynamics, particle simulations routinely produce datasets of billions to trillions of records per snapshot~\cite{habib2016hacc}, with individual snapshots reaching hundreds of terabytes and cumulative data products reaching the exabyte scale~\cite{frontiere2025cosmological,tchipev2019twetris}. Retaining and transferring these datasets at full precision is increasingly impractical because lossless compression offers limited relief, yielding only about 2:1 on floating-point data~\cite{zhao2021optimizing}. Error-bounded lossy compression has therefore become a widely used data-reduction technique across scientific domains~\cite{jin2022accelerating,ren2026ffcz,zhang2025lcp,liang2020toward,liang2022toward}, reducing data volume by an order of magnitude or more under user-controlled pointwise error bounds.

However, many existing lossy compression methods are not applicable to particle data.
Regular-grid compressors such as SZ3~\cite{liang2022sz3} and ZFP~\cite{lindstrom2014fixed} assume that neighboring array indices correspond to nearby spatial locations and therefore yield low compression ratios on irregular particle data because they do not exploit the spatial coherence of particles. Point-cloud codecs such as Google Draco~\cite{draco}, MPEG G-PCC~\cite{gpcc}, and MPEG V-PCC~\cite{vpcc} exploit spatial geometry, but their lossy modes may map multiple points to the same reconstructed location and merge duplicates. Such merging may be acceptable for interchangeable samples of a surface or scene, but in particle simulations, each particle is an identifiable entity with physical attributes; changing the particle count breaks the correspondence needed to align fields, track trajectories, and analyze interactions.

While more recent compressors like LCP~\cite{zhang2025lcp} and MDZ~\cite{zhao2022mdz} are specifically tailored for particle data, they still focus exclusively on bounding coordinate reconstruction errors. Many downstream analyses instead query clustering structures derived from those coordinates.
A prominent example is single-linkage clustering, used for Friends-of-Friends (FoF) halo finding in cosmology~\cite{more2011overdensity,rodriguez2020combining}, coherent structure detection in fluid dynamics~\cite{monchaux2010preferential,west2024clustering,colanera2025extended}, and analyses of ion cluster and protein aggregation in molecular simulations~\cite{france2019correlations,li2019dissipative}. Even small coordinate errors near the linking threshold can break true links, create false links, and thereby merge or split entire clusters.

Existing compressors can preserve cluster membership only indirectly in three ways, each sacrificing storage efficiency or data utility. First, imposing sufficiently strict global error bounds to prevent cluster merges or splits forces all coordinates to near-lossless precision, even though perturbations to most coordinates cannot change cluster membership, and therefore yields limited compression ratios. Our empirical tests on an HACC dataset~\cite{sdrbench} with over one billion 3D particles demonstrate that SZ3 and LCP require a relative error bound around $10^{-8}$ to preserve cluster membership and attain compression ratios of only 1.42 and 1.75, respectively. Second, storing cluster membership as a separate integer array of particle indices alongside the compressed coordinates decouples membership from coordinate error but adds substantial metadata.
While cosmological analyses often ignore clusters below a specified particle-count threshold, the membership indices of approximately $0.1N$-$0.2N$ particles still require storage, where $N$ is the number of particles. For example, the HACC dataset~\cite{sdrbench} with 1.07 billion particles has 113.8 and 168.7 million particles within clusters meeting typical size thresholds of 100~\cite{lukic2007halo,warren2006precision} and 20~\cite{springel2005simulations}, respectively; storing these indices as 4-byte integers adds 7.1\% and 10.5\% relative to the original storage. Moreover, this metadata does not improve the fidelity of compressed coordinates for other downstream analyses. Third, reordering particles by halo encodes membership implicitly in the data layout with small indexing overhead, but breaks stable particle indices used to align coordinates with other fields and track particles over time; storing a recovery permutation largely eliminates the resulting storage benefit. These limitations necessitate methods that achieve high cluster fidelity while maintaining good compression ratios.

We propose a novel correction-based method that preserves single-linkage cluster membership under lossy compression with a user-defined error bound $\xi$ and distance threshold $b$. The method is applied in situ during compression and is compatible with any off-the-shelf lossy compressor (referred to as the \textit{base compressor}). We first use a spatial cell partition to identify near-threshold particle pairs whose links are vulnerable under error bound $\xi$. Since membership queries depend on connected components rather than the complete proximity graph, we then introduce vulnerable-pair, safe-component, and halo-forest modes that progressively reduce the enforced constraints. Next, we formulate cluster preservation as a constrained optimization problem, where the loss function penalizes violations of pairwise distance constraints. We apply projected gradient descent to enforce the selected constraints while retaining the pointwise error bound of the base compressor, thereby preserving the original connected components upon convergence. Each correction stream targets one threshold $b$, matching scientific workflows that commonly fix the linking length in cosmology~\cite{more2011overdensity,habib2016hacc,heitmann2019hacc} and molecular dynamics~\cite{abramyan2016cluster,brielle2020quantitative,wedekind2007best}. Supporting multiple thresholds is possible by taking the union of their selected pair sets, but the additional constraints and metadata reduce compression ratio and throughput. To ensure scalability, we
provide CPU, GPU, and distributed-memory MPI implementations.

We evaluate our method on large-scale cosmology, molecular dynamics, and fluid dynamics datasets using SZ3, ZFP, cuSZp2, Draco, and LCP as base compressors. At full convergence, our method exactly preserves cluster membership; under a fixed iteration budget, it improves clustering and the accuracy of downstream statistics. We use the adjusted Rand index (ARI) and intersection over union (IoU) to measure the global fidelity of pairwise same-halo and halo-membership query results, respectively; the halo mass function (HMF), an important scientific analysis curve, is also used. Our method preserves cluster membership while improving compression ratio by up to $4\times$ relative to the same base compressors configured with error bounds sufficiently tight to preserve cluster membership. The GPU solver achieves up to $93\times$ end-to-end speedup over the CPU baseline, and the throughput is competitive with the evaluated base compressors, confirming that the correction step integrates into high-throughput simulation pipelines without introducing a computational bottleneck. 
Our contributions include:

\begin{itemize}
   \item \textbf{A constraint hierarchy for cluster preservation}: We show that single-linkage cluster-membership queries depend on connected components rather than the complete set of proximity links, and derive three progressively smaller constraint sets.
   \item \textbf{A correction algorithm}: We propose a correction algorithm that preserves single-linkage cluster membership for particle data while satisfying user-defined coordinate error bounds.
   \item \textbf{GPU and distributed parallelism}: We design and implement a distributed, data-parallel solver that scales across multi-node CPU and GPU clusters via MPI.
   \item \textbf{Comprehensive evaluation}: We validate the method across multiple scientific datasets and state-of-the-art base compressors using query-oriented clustering metrics.
\end{itemize}

%% file: 2.related.tex
\section{Related Work and Background}
\label{sec:related}
We review how prior compressors exploit the organization of regular grids, point clouds, and scientific particles, and whether these compressors preserve coordinates or derived features. We then relate cluster preservation to query-preserving data reduction in data management and define the single-linkage clustering queries considered in this work.

\subsection{Error-bounded Lossy Compression}
Error-bounded lossy compression reduces data volume while ensuring that the pointwise difference between original and reconstructed values remains within a user-specified bound.

\textbf{Regular-grid compressors} are commonly categorized by their reconstruction primitives~\cite{di2025survey}. Prediction-based methods, such as the SZ family, including SZ3~\cite{liang2022sz3} and cuSZp2~\cite{huang2024cuszp2}, use predictors (e.g., the Lorenzo predictor~\cite{ibarria2003out}) to estimate data values based on their neighbors and quantize the residual error to satisfy strict $L_\infty$ guarantees. In contrast, transform-based methods, such as ZFP~\cite{lindstrom2014fixed}, employ block-based transforms to map data into sparse coefficients that are more amenable to compression. While these tools achieve high compression ratios for smooth, structured fields, they rely on the implicit spatial locality of grid indices. For irregular data, where index-level adjacency does not correspond to spatial proximity, these methods yield significantly degraded compression ratios.

\textbf{Point-cloud compressors} such as Google Draco~\cite{draco}, MPEG G-PCC~\cite{gpcc}, and V-PCC~\cite{vpcc} exploit geometric structure through spatial partitioning, coordinate quantization, or projection-based representations. These codecs target sampled surfaces and scenes, where exact point identity and multiplicity are not always required. Their lossy modes may map multiple input points to the same reconstructed location, which is unsuitable when every simulation particle and its associated attributes must remain identifiable.

\textbf{Particle-specific compressors} preserve particle identity while exploiting locality in scientific particle data. Domain-specific methods like MDZ~\cite{zhao2022mdz} exploit unique spatial patterns (e.g., zigzag or stairwise) and high temporal similarity in molecular dynamics trajectories. More recent advancements, such as LCP~\cite{zhang2025lcp}, use block-wise decomposition to improve data-parallel processing. However, their guarantees remain pointwise: bounded coordinate reconstruction error does not guarantee preservation of derived clusters.

\subsection{Feature-preserving lossy compression}

Feature-preserving lossy compression goes beyond pointwise reconstruction guarantees by controlling specific derived properties, often called \textit{quantities of interest} (QoIs), computed from decompressed data. These QoIs typically include structural, statistical, physical, and topological properties used in downstream scientific analyses~\cite{di2025survey}. For example, QPET~\cite{liu2025qpet} is designed to maintain the quantile distribution of the data.

\textbf{Correction-based approaches}, also called edit- or augmentation-based approaches, preserve QoIs without modifying the base compression algorithm. These methods refine the decompressed output by applying ``edits'' to restore specific features while strictly adhering to the original pointwise error bounds. For example, to preserve power spectra, FFCz~\cite{ren2026ffcz} derives edits by alternating projections between spatial and frequency constraints. To preserve scalar field topology, MSz~\cite{li2024msz} edits decompressed data to maintain Morse–Smale segmentations, while Gorski et al.~\cite{gorski2025general} focus on contour tree accuracy.
However, these methods primarily target regular-grid scalar fields and do not preserve connectivity-based structures in particle data.

\subsection{Query-preserving data reduction in data management}

Data management research has long studied compact representations that preserve or approximate downstream query results. Approximate query processing (AQP) uses samples, histograms, wavelets, and sketches to estimate aggregate queries such as counts, sums, and quantiles~\cite{cormode2004improved,garofalakis2001approximate,garofalakis2002wavelet,wu2009distributed,wu2010continuous}. Semantic and model-based compression instead exploits correlations and learned patterns to reconstruct approximate attribute values from compact relational representations~\cite{jagadish1999semantic,babu2001spartan,jagadish2004itcompress}. These approaches provide guarantees on query estimates or reconstructed values but do not jointly reconstruct every particle within a pointwise coordinate bound or preserve the exact partition induced by proximity connectivity.

FoF membership is a connected-component query over a graph whose edges are pairwise-distance predicates. Because connectivity is transitive, flipping one near-threshold predicate can change the component membership of many particles. Our method therefore provides a different guarantee from AQP and semantic compression: it retains the full particle data under a coordinate error bound while preserving connected-component answers exactly upon convergence. Identifying near-threshold particle pairs is closely related to a spatial distance self-join~\cite{brinkhoff1993efficient}; however, our objective is not to preserve every join result, but to enforce a sufficient set of constraints to keep the connected components unchanged. Table~\ref{tab:reduction_category} summarizes the representative approaches.

\begin{table}[!ht]
      \centering
      \footnotesize
      \caption{Comparison of representative data-reduction approaches.
      A pointwise bound constrains reconstructed values, whereas a downstream
      guarantee constrains a derived analysis or query result. Particle identity indicates whether each input particle has a distinct reconstructed counterpart.}
      \begin{tabular}{c|c|c|c|c}
      \toprule
          Method &
          \makecell{Target data} &
          \makecell{Pointwise\\bound?} & \makecell{Particle\\identity?} &
          \makecell{Downstream\\guarantee} \\
      \midrule
          \makecell{SZ3, ZFP,\\cuSZp2} &
          regular grids &
          Y &
          N/A &
          none \\
      \hline
          \makecell{Draco, G-PCC,\\V-PCC} &
          point clouds &
          \makecell{codec-\\dependent} &
          \makecell{not\\guaranteed} &
          none \\
      \hline
          \makecell{MDZ, LCP} &
          particles &
          Y &
          Y &
          none \\
      \hline
          \makecell{FFCz, QPET,\\MSz} &
          \makecell{regular grids} &
          Y &
          N/A &
          \makecell{spectral/statistical/\\topological QoIs} \\
      \hline
          AQP &
          \makecell{relations/\\streams} &
          N/A &
          N/A &
          \makecell{approximate\\aggregate queries} \\
      \hline
          \makecell{Semantic\\compression} &
          relations &
          \makecell{method-\\dependent} &
          N/A &
          \makecell{approximate\\attribute values} \\
      \hline
          Ours &
          particles &
          Y &
          Y &
          \makecell{cluster membership} \\
      \bottomrule
      \end{tabular}
      \label{tab:reduction_category}
  \end{table}

\subsection{Clustering-query background}
\label{sec:fof}
\begin{figure}[!ht]
    \centering
    \includegraphics[width=\linewidth]{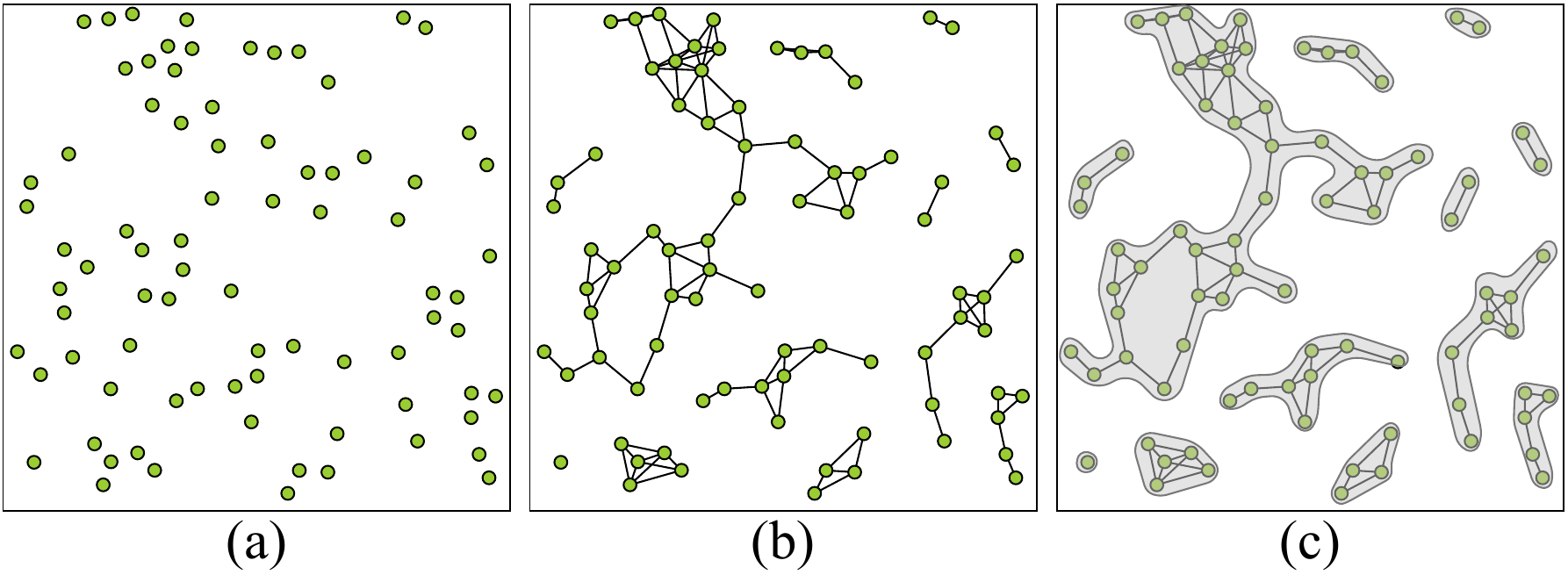}
    \caption{Illustration of the FoF clustering algorithm. (a) An initial distribution of particles in space. (b) Edge construction for all particle pairs with a Euclidean distance $d(p_i, p_j) \le b$, where $b$ is the linking length. (c) The resulting connected components, which define the final clusters (halos).}
    \label{fig:FoF_alg}
\end{figure}

Our preservation target is the partition produced by single-linkage clustering, known as the friends-of-friends (FoF) algorithm in cosmology. FoF is a foundational tool across numerous scientific domains~\cite{more2011overdensity,france2019correlations,monchaux2010preferential}, defining clusters based on a proximity rule: for a set of $N$ particles $P=\{p_n\}_{n=0}^{N-1}$, the FoF algorithm identifies clusters based on a proximity threshold called the \textit{linking length} denoted by $b$. We define a connectivity graph $G=(P,E)$ where an edge $(p_i, p_j)\in E$ exists if and only if the Euclidean distance of the two particles $p_i$ and $p_j$ satisfies $d(p_i,p_j)\leq b$. Each FoF cluster (referred to as a ``halo'' in cosmology~\cite{roy2014pfof}) corresponds to a connected component of $G$ (Fig.~\ref{fig:FoF_alg}). This partition supports two downstream queries evaluated in this work: whether two particles belong to the same cluster, and which cluster a given particle belongs to.

In practice, $b$ is determined by a \textit{linking parameter} $\eta$ and the \textit{mean particle separation} $\Delta_p$. For a dataset in $d$-dimensional space with a total volume $V_{vol}$, $b$ is defined as:
\begin{equation*}
b = \eta\Delta_p, \quad \text{where} \quad \Delta_p = \left( \frac{V_{vol}}{N} \right)^{1/d}
\end{equation*}
In cosmological simulations, $\eta$ is typically set to $0.2$ or $0.168$~\cite{more2011overdensity,habib2016hacc,heitmann2019hacc}. This formulation ensures that the clustering threshold scales appropriately with the global density.

\subsubsection{Single-linkage Clustering Implementation}
Efficiently computing FoF clusters for large-scale datasets requires specialized spatial indexing and parallel algorithms to avoid the $O(N^2)$ complexity of a naive pairwise search~\cite{springel2001gadget,habib2016hacc}.
A widely adopted strategy involves \textit{spatial partitioning and cell linking}, where the simulation volume is partitioned into a uniform grid of cells~\cite{habib2016hacc}. By setting the cell side length $w\geq b$, the search for any particle pair $(p_i,p_j)$ such that $d(p_i,p_j)\leq b$ is restricted to pairs located within the same cell or two adjacent cells. In parallel environments, each thread typically processes a unique cell and evaluates candidate pairs within that cell and its forward neighboring cells. By only checking neighbors in one direction (e.g., 4 specific neighbors in 2D or 13 in 3D), the algorithm avoids redundant distance computations for the same pair of cells. This cell-linking strategy reduces the complexity of edge identification to approximately $O(N \cdot \bar{\rho})$, where $\bar{\rho}$ represents the average number of particles per cell.

\textbf{Parallel clustering with union-find}. Connected components are typically computed using the union-find (or disjoint set union) data structure~\cite{tarjan1975efficiency,woodring2011situ}. Each particle starts as a separate set; for every pair $(p_i, p_j)$ satisfying the distance constraint, a \texttt{union} operation merges their respective sets. To handle massive concurrency, modern implementations often apply Rem-Lock-Free algorithms~\cite{patwary2010experiments} or pointer jumping techniques~\cite{vishkin1982log} with atomic operations across thread or process boundaries without the overhead of traditional locking mechanisms. The final state of the union-find structure provides the cluster ID for each particle, representing the clustering outcome that our method aims to preserve.

\subsubsection{Single-linkage Clustering Applications}

In cosmology, single-linkage clustering is a standard for halo finding, used to identify dark matter structures in N-body simulations~\cite{more2011overdensity,rodriguez2020combining}. Similarly, in molecular dynamics, it is used to identify ion-pair clusters and characterize molecular self-assembly processes~\cite{france2019correlations,li2019dissipative}. It is also critical in fluid dynamics for detecting coherent structures in turbulent flows~\cite{monchaux2010preferential,west2024clustering,colanera2025extended}. In these contexts, the resulting clusters, including mass functions and spatial distributions, form the essential basis for high-level scientific inference.

The \textit{halo mass function} (HMF), denoted by $dn/d\log M$, represents the number density of dark matter halos per logarithmic mass interval and serves as a critical downstream metric for validating cosmological simulations~\cite{reed2007halo}. The HMF is a fundamental probe of the $\Lambda$ cold dark matter ($\Lambda$CDM) model, as its shape and evolution are directly governed by the growth of cosmic structures and the underlying dark matter density field~\cite{lukic2007halo,tinker2008toward,zentner2007excursion}. For each identified halo $i$, we first calculate its total mass $M_i$; assuming uniform particle mass, $M_i \propto N_i$, where $N_i$ is the number of constituent particles in halo $i$. We then partition these masses into $B$ equal-width logarithmic bins and normalize the resulting counts by the total simulation volume and the logarithmic bin width.

Because the HMF is computed from FoF component sizes, membership errors change halo masses and bin counts and can lead to a fundamental mischaracterization of the universe, such as underestimating the total amount of matter or miscalculating how ``clumpy'' the cosmic structure is.

%% file: 3.method.tex
\section{Methodology}
\label{sec:method}

This section formulates the problem of preserving FoF clusters as a constrained optimization problem, introduces three constraint-selection modes, describes our iterative refinement algorithm, and details its implementation for single-node GPU and multi-node environments. While our current implementation assumes a non-periodic domain, periodic support does not change the core formulation but requires minimum-image distances, wrapped cell neighborhoods, and periodic ghost exchange; we leave this extension to future work.

\subsection{Constrained optimization problem}
\label{sec:optimization_problem}
We formulate the preservation of FoF clusters as a constrained optimization problem, where the objective is to minimize the cumulative distance overflow of violated pairs subject to the global pointwise error bound. Let the original 3D particle data be $P=\{p_n\}_{n=0}^{N-1}$ with coordinates $(x_n, y_n, z_n)$, and the corresponding decompressed data from a base compressor be $\hat{P}=\{\hat{p}_n\}_{n=0}^{N-1}$ with coordinates $(\hat{x}_n, \hat{y}_n, \hat{z}_n)$. Given an absolute error bound $\xi$, the reconstructed coordinates satisfy $|\hat{x}_n-x_n|\leq\xi$, $|\hat{y}_n-y_n|\leq\xi$, and $|\hat{z}_n-z_n|\leq\xi$ for all $n$, implying a maximum displacement of $\sqrt{3}\xi$ for any single particle. Consequently, the Euclidean distance $d(\hat{p}_i,\hat{p}_j)$ between two arbitrary decompressed particles can deviate from its original value $d(p_i,p_j)$ by at most $2\sqrt{3}\xi$. This $2\sqrt{3}\xi$ bound represents the worst-case scenario where two particles are displaced in opposite directions along the cube diagonals. While the average distance deviation is significantly lower in practice, this conservative radius is necessary to identify potential connectivity changes.

Under linking length $b$, an original link with distance in $(b-2\sqrt{3}\xi,b]$ can break (a \textit{potentially-broken link}), whereas a link can be created for an originally non-linked pair with distance in $(b,b+2\sqrt{3}\xi]$ (a \textit{potentially-created link}). We call their union the set of \textit{vulnerable pairs}, $V=\{(i,j):d_{ij}\in(b-2\sqrt{3}\xi,b+2\sqrt{3}\xi]\}$. Each set of pairs whose preservation can guarantee invariant FoF clusters is a subset of $V$; we call such a set to be a set of \textit{active pairs} (denoted by $A$). The set of \textit{editable particle} $E$ contains the endpoints of pairs in $A$, as modifying only their positions in the decompressed data is sufficient to restore the correct clusters.

To preserve the clusters, we define a loss function $\mathcal{L}$ that penalizes distance deviations for pairs whose connectivity has been altered. The loss is formulated as:
\begin{equation}
\mathcal{L}(\hat{P})=\sum_{(i,j)\in A:d_{i,j}\leq b<\hat{d}_{i,j}} (\hat{d}_{i,j}-b)^2+\sum_{(i,j)\in A:\hat{d}_{i,j}\leq b<d_{i,j}} (b-\hat{d}_{i,j})^2,
\label{eq:loss}
\end{equation}
where $d_{i,j} = d(p_i, p_j)$ and $\hat{d}_{i,j} = d(\hat{p}_i, \hat{p}_j)$. The first term penalizes currently broken active links, while the second term penalizes currently created active links. Only active pairs contribute to this loss, and only editable particles are involved in the optimization. We then formulate the following constrained optimization problem to refine the particle positions:
\begin{align}\label{eq:optimization}
\min_{\hat{p}_n:n\in E} \quad & \mathcal{L}(\hat{P}) \\ 
\text{s.t.} \quad & |\hat{x}_n-x_n| \leq \xi, 
\quad|\hat{y}_n-y_n| \leq \xi,\quad|\hat{z}_n-z_n| \leq \xi,\quad \forall n\in E. \nonumber
\end{align}

\begin{algorithm}[!ht]
\footnotesize
\caption{FoF Cluster Preserving Correction}
\label{alg:main}
\begin{algorithmic}
\REQUIRE Original positions $P$, base-compressed positions $\hat{P}^{(0)}$,
         error bound $\xi$, linking length $b$, bit depth $m$, learning rate $\alpha$,
         max iterations $T_{\max}$, convergence threshold $\epsilon_\mathcal{L}$, mode
\ENSURE Compressed bitstream of FoF cluster-preserved edits
\STATE $\epsilon_q \leftarrow 2\xi / (2^m - 1)$
\STATE $\xi' \leftarrow \xi(1 - 2^{-m})$
\STATE $V \leftarrow \{(i,j) : d_{ij} \in (b - 2\sqrt{3}\xi, b + 2\sqrt{3}\xi]\}$
\STATE $A \leftarrow \mathrm{SelectConstraints}(V, \mathrm{mode})$
\STATE $E \leftarrow endpoints(A)$
\STATE $\hat{P} \leftarrow \hat{P}^{(0)}$
\FOR{$t = 1$ \TO $T_{\max}$}
    \IF{$\mathcal{L}_{\text{tight}}(\hat{P}) \leq \epsilon_\mathcal{L}$}
        \STATE \textbf{break}
    \ENDIF
    \STATE $g \leftarrow \nabla_{\hat{P}}\,\mathcal{L}_{\text{tight}}(\hat{P})$
    \STATE $\hat{P} \leftarrow \hat{P} - \alpha\, g$
    \STATE $\hat{P} \leftarrow \mathrm{proj}_{\mathcal{B}(\xi')}(\hat{P})$
\ENDFOR
\STATE $\Delta \leftarrow \hat{P} - \hat{P}^{(0)}$
\STATE \textit{flags} $\leftarrow$ $N$-bit mask of particles with nonzero edit vectors in $\Delta$
\STATE \textit{edits} $\leftarrow$ three-coordinate edits of flagged particles, quantized to $m$ bits
\RETURN Huffman+ZSTD(\textit{flags}, \textit{edits})
\end{algorithmic}
\end{algorithm}

\begin{figure}[!ht]
    \centering
    \includegraphics[width=\linewidth]{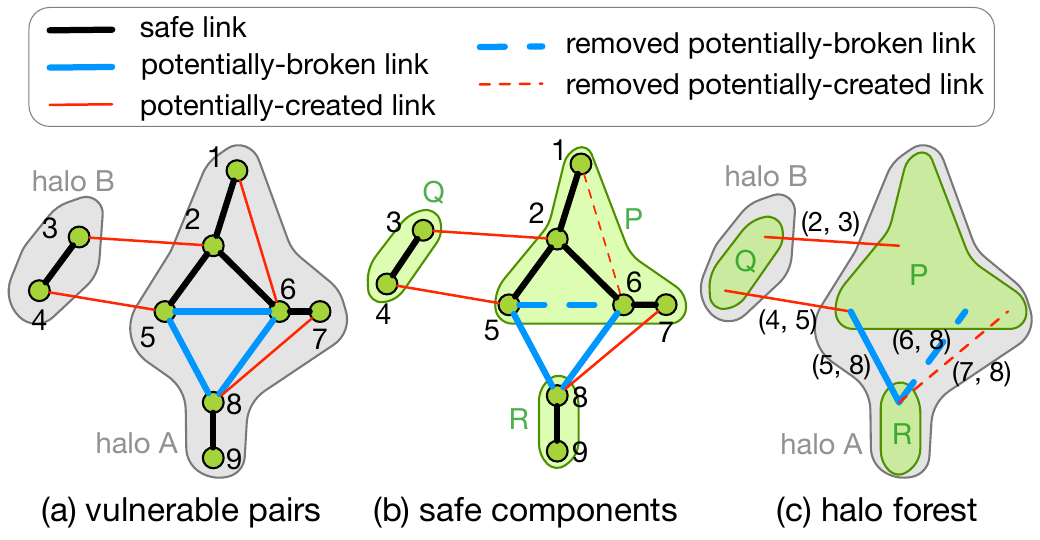}
    \caption{Illustration of the three constraint-selection modes. (a) Vulnerable-pair mode keeps all vulnerable pairs, including potentially-broken and potentially-created links. (b) Safe-component mode removes vulnerable links whose endpoints are connected by a path of safe links, i.e., lie within the same safe component, such as P, Q, and R. (c) Halo-forest mode contracts each safe component into a supernode, additionally removes potentially created links (e.g., link (7, 8)) between supernodes within the same original halo, and preserves only a spanning forest of potentially-broken links (i.e., link (5, 8)) among supernodes within each halo.}
    \label{fig:3modes}
\end{figure}

\subsection{Constraint-selection modes}
\label{sec:modes}
Enforcing every vulnerable pair (i.e., setting $A=V$) preserves all near-threshold link predicates, but this is stronger than many downstream queries require. A cluster-membership query does not require every individual link; it depends only on the connected components induced by those links. We therefore introduce three modes that select progressively smaller active-pair sets $A$, each preserving a coarser clustering query at lower correction cost (Fig.~\ref{fig:3modes}).

\noindent\textbf{Mode 1: vulnerable-pair (VP)}. We set the active set $A_{VP}=V$, enforcing every vulnerable pair. This preserves the existence of every near-threshold link, and hence the exact answer to a link-existence query. It gives the strongest guarantee and the largest constraint set.

\noindent\textbf{Mode 2: safe-component (SC)}. We call a particle pair $(p_i,p_j)$ a \textit{safe link} if their distance $d_{ij}\le b-2\sqrt{3}\xi$. Since compression perturbs any distance by at most $2\sqrt{3}\xi$, a safe link is guaranteed to survive after base compression without correction. The connected components induced by safe links are called \textit{safe components}, each guaranteed to remain a single connected subcluster in the reconstructed data. In this mode, we discard any vulnerable pair whose endpoints already lie in the same safe component, keeping only inter-component pairs.

\noindent\textbf{Mode 3: halo-forest (HF)}. Membership preservation does not require enforcing \textit{every} inter-component pair. In addition to SC mode, this mode contracts each safe component into a supernode and computes the original halos, i.e., the connected components of the graph on supernodes whose edges are the potentially-broken links. Within each halo, we discard the potentially-created links, while potentially-created links between different original halos remain active to prevent merges. We keep only a subset of the inter-supernode potentially-broken links that forms a \textit{spanning forest} of all supernodes in each halo. The active pair sets $A$ of the three modes form a hierarchy
$A_{HF}\subseteq A_{SC}\subseteq A_{VP}=V$.

\subsection{Projected gradient descent}

We propose a projected gradient descent (PGD) approach to iteratively refine particle coordinates, where each iteration performs a gradient descent to minimize the violation loss, followed by a projection of the updated coordinates back onto the box constraints defined by $\xi$, as shown in Alg.~\ref{alg:main}.

\textbf{Identifying safe links and vulnerable pairs}. For a given linking length $b$ and error bound $\xi$, we identify safe links (for modes SC and HF only) and vulnerable pairs using a spatial partitioning and cell-linking strategy, as detailed in Section~\ref{sec:fof}. Specifically, we set the cell side length to $w\geq b+2\sqrt{3}\xi$, representing the maximum possible distance for a vulnerable pair. To maintain computational efficiency and avoid an overabundance of empty cells in sparse regions, we constrain the total number of cells to be no greater than the number of particles $N$. We then iterate through all particle pairs within the same cell or adjacent cells, using a directional search to avoid redundant cell pairs, to identify the set of safe links and vulnerable pairs.

\textbf{Active pair selection} (for modes SC and HF only). In SC mode, we compute the safe components induced by safe links via union-find and discard all vulnerable pairs within the same safe component. In HF mode, we additionally compute the FoF halos induced by potentially-broken links between contracted safe components via another round of union-find, discard potentially-created links within a halo, and preserve only a spanning tree of potentially-broken links for each halo. After selecting active pairs, we derive their constituent editable particles.

\textbf{Gradient descent and projection}. The optimization begins by calculating the initial loss and terminates immediately if it is below a predefined tolerance. Otherwise, each iteration first updates the editable-particle coordinates by gradient descent and then projects any out-of-bound coordinates onto the box defined by $\xi$, enforcing Eq.~\eqref{eq:optimization}. The process repeats until the loss satisfies the convergence criterion or the maximum iteration count is reached.

\textbf{Compaction, quantization, and lossless compression}.
Upon termination of the optimization, the coordinate shifts are recorded as ``edits'' and decomposed into two components: \textit{flags} and \textit{compact edits}. The flags consist of an $N$-bit mask indicating which particles were modified and are packed into 8-bit integers for efficient storage. The compact edits store the three coordinate values for each flagged particle, at most $3|E|$ values. These edits undergo uniform quantization into $2^m$ intervals, where $m$ denotes the bit depth.

To ensure that the quantized edits strictly satisfy the box constraints in Eq.~\eqref{eq:optimization}, we apply a safety margin by shrinking the error bound to $\xi' = \xi(1 - 2^{-m})$. Furthermore, to ensure the restored clusters are robust against the maximum possible quantization error $\epsilon_q=\frac{2\xi}{2^m-1}$, in practice we optimize a tightened loss $\mathcal{L}_{\text{tight}}$ in place of $\mathcal{L}$ in Eq.~\eqref{eq:optimization} during the PGD:
\begin{align}
\mathcal{L}_{\text{tight}}(\hat{P})=&\sum_{(i,j)\in A:d_{i,j}\leq b, \hat{d}_{i,j}>b-2\sqrt{3}\epsilon_q} (\hat{d}_{i,j}-b+2\sqrt{3}\epsilon_q)^2\\
&+\sum_{(i,j)\in A:d_{i,j}>b,\hat{d}_{i,j}\leq b+2\sqrt{3}\epsilon_q} (b+2\sqrt{3}\epsilon_q-\hat{d}_{i,j})^2.\nonumber
\label{eq:tight_loss}
\end{align}
This tightened loss moves each corrected pair far enough from $b$ to absorb the error introduced when its edits are quantized and decoded. Therefore, if $\mathcal{L}_{\text{tight}}=0$ before edit quantization, applying the decoded edits keeps the corresponding untightened loss $\mathcal{L}$ at zero. To assess sensitivity to $m$, we evaluate $m \in \{8, 16, 32\}$ on a molecular dynamics dataset, EXAALT, as a representative case; zero violations are re-introduced by quantization in all cases, confirming robustness across a wide range of bit-depths. Among these, $m=16$ achieves the best compression ratio due to the tradeoff between edit precision and encoding overhead, and is used as the default in all experiments. Finally, the flags and quantized edits are compressed using Huffman coding~\cite{huffman1952method} followed by ZSTD~\cite{zstd} to further reduce the storage.

\textbf{Reconstruction of the edited decompressed data}. To reconstruct the data, we first apply ZSTD decompression and Huffman decoding to retrieve the particle mask and quantized edits. The compact edits are then dequantized and mapped to the flagged particles as three-coordinate edit vectors. The final coordinates are obtained by adding these edits to the corresponding output positions of the base compressor.

\textbf{Convergence analysis}. In practice the algorithm projects onto the tighter box $\mathcal{B}(\xi')$ with $\xi'=\xi(1-2^{-m})<\xi$ to absorb quantization error; the original positions $P$ lie at the center of $\mathcal{B}(\xi)$ and therefore also within $\mathcal{B}(\xi')$, so $P$ is feasible within $\mathcal{B}(\xi')$; moreover, since $2\sqrt{3}\epsilon_q\ll\xi$, nearby positions achieve $\mathcal{L}_{\text{tight}} = 0$, confirming the global minimum is zero. Although the loss is non-convex due to the false-link term, both components are $C^1$ with Lipschitz-continuous gradients since each term has the form $[\max(0, g)]^2$ where $g$ is smooth~\cite{beck2017first}, enabling monotonic loss decrease and $O(1/T)$ convergence to a stationary point under PGD. Two structural properties promote convergence to the global minimum in practice. First, potentially-created links among vulnerable pairs are rare, especially at small error bounds, so the loss is dominated by its convex potentially-broken links component in most configurations. Second, the vulnerable-pair graph decomposes into many small independent components optimized separately; within each component, the geometric conditions required for gradient cancellation between conflicting pairs are extremely unlikely to be satisfied simultaneously. Together, these properties explain why PGD reaches $\mathcal{L} = 0$ in all tested configurations when run to convergence without iteration cap, as confirmed empirically in Section~\ref{sec:eval}, and the iteration budget scales with the number of violations rather than total particle count. The same analysis applies to $\mathcal{L}_{\text{tight}}$, whose global minimum of zero is achievable since the additional margin $2\sqrt{3}\epsilon_q \ll \xi$ is easily accommodated within the box constraints. In practice, we use Adam~\cite{kingma2014adam} for faster convergence; the theoretical guarantees above apply to vanilla PGD, and empirical convergence is confirmed in Section~\ref{sec:eval}.

\subsection{Single-GPU Acceleration}

Our GPU implementation parallelizes all stages. During spatial partitioning, each thread maps one particle to its grid cell and atomically increments per-cell counts in global memory; a device-wide exclusive prefix sum (via the CUB library~\cite{cub}) then converts these counts into cell-start offsets, after which a scatter kernel places each particle into a cell-sorted array using atomic index reservations. This cell-sorted layout ensures that subsequent neighbor searches access spatially co-located particles, improving global memory coalescing.

Vulnerable-pair detection scans the cell-pair list in two passes. The first counts qualifying links and, for SC and HF modes, builds safe components (and original-halo components for HF) with GPU union-find. Each thread processes a strided subset of particle pairs and uses \texttt{atomicCAS} to hook the higher-index root under the lower-index root, retrying after a race. Then a subsequent kernel flattens parent chains by path halving. HF applies the same operation to safe-component roots and retains a potentially-broken inter-component link only when its union succeeds, thereby constructing a spanning forest. SC and HF then either generate active links directly from the roots or write and filter all vulnerable links, where the latter path uses CUB \texttt{DeviceSelect::Flagged} for compaction. Finally, the active pairs are converted into a dense lookup map and compact list of editable particles.

In the PGD phase, threads compute per-pair loss and gradients, which are summed through a shared-memory reduction within each block and accumulated into a global buffer via a single atomic addition per block. Gradients are computed in a per-pair fashion and accumulated into per-coordinate buffers through global-memory atomic additions indexed by the direct-address map. The gradient update and box projection are then applied embarrassingly in parallel, with one thread per editable particle.

\subsection{Inter-node distributed-Memory Scaling}
To scale the algorithm to cosmological volumes exceeding a single GPU's memory, we provide a distributed-memory implementation using the Message Passing Interface (MPI). The communication overhead consists of three phases. First, \textbf{bounding box exchange} pads each rank's bounding box by a ghost zone of width $\delta = b + 2\sqrt{d}\xi$ in $d$ dimensions and uses \texttt{MPI\_Allgather} to exchange bounding-box metadata; two ranks are neighbors if their padded volumes overlap. Second, \textbf{ghost exchange} transfers particle coordinates from neighboring ranks via nonblocking \texttt{MPI\_Isend} and \texttt{MPI\_Irecv}. We overlap these transfers with allocation of the extended particle buffer and copying of local particles, synchronizing only before spatial partitioning. Each rank runs union-find independently on its extended local-plus-ghost array. GPU ranks use atomic minimum-root hooking followed by root flattening, whereas CPU ranks use minimum-root linking with path halving. The pair scan excludes ghost--ghost pairs but includes local--local and local--ghost pairs to expose boundary constraints. Because the roots are not reconciled globally, SC and HF retain all vulnerable local--ghost pairs rather than pruning them by local component labels, avoiding a distributed union-find. Third, \textbf{global convergence check} uses an \texttt{MPI\_Allreduce} of the loss at every iteration to synchronize the stopping decision. Only rank-local particles are edited and written to the final edit stream, while ghost particles serve only to expose boundary constraints during local computation.

\subsection{Complexity and scalability analysis}

Let $C$ be the number of nonempty neighboring cell pairs and $K$ the number of candidate particle pairs examined within those cell pairs. Spatial partitioning costs $O(N)$ work plus prefix-scan overhead. Vulnerable-pair detection costs $O(K)$ work, where $K=O(N\rho)$ under bounded cell occupancy and average local density $\rho$. Mode VP has $A=V$; SC and HF add union-find and filtering costs near-linear in the number of safe/vulnerable links but reduce $|A|$.

Each PGD iteration costs $O(|A|+|E|)$ work, because active pairs contribute loss and gradient terms, and editable particles are updated and projected independently. Thus the total optimization work is $O(T(|A|+|E|))$, where $T$ is the number of iterations. Since $|A|$ decreases from VP to SC to HF, the later modes reduce both PGD cost and edit storage, at the expense of extra union-find and filtering work during preprocessing.

In the distributed setting, each rank communicates only ghost particles within a boundary layer of width $\delta=b+2\sqrt{3}\xi$. For approximately uniform 3D decompositions, the ghost volume scales with the subdomain surface area, giving communication proportional to $O((N/R)^{2/3})$ per rank up to density and boundary-width constants, while local computation scales with the rank-local particles and active links, where $R$ represents the number of ranks.

%% file: 4.evaluation.tex
\section{Evaluation}
\label{sec:eval}
We evaluate (1) storage efficiency and coordinate distortion, (2) fidelity of cluster-membership and aggregate queries, and (3) single-GPU and distributed-memory performance.

\subsection{Evaluation scheme}

\textbf{Baselines and base compressors}. We evaluate our method using SZ3~\cite{liang2022sz3}, ZFP~\cite{lindstrom2014fixed}, cuSZp2~\cite{huang2024cuszp2}, Google Draco~\cite{draco}, and LCP~\cite{zhang2025lcp} as representative baselines and base compressors. SZ3 and ZFP are state-of-the-art general-purpose lossy compressors that provide strict pointwise error control and are widely used across diverse scientific applications. cuSZp2 is a GPU-accelerated compression technique based on SZ3. Draco and LCP are specifically designed for particle data. Furthermore, we implement a Morton-order baseline that operates post-spatial partitioning; this method traverses particles within each cell according to their Morton codes, applies a first-order predictor and subsequent error quantization to provide a reference for spatial-locality-based compression.

\textbf{Metrics}. We compare these baselines against the same base compressors corrected by our method using quantitative metrics, including compression ratio, rate distortion, and throughput. To evaluate FoF cluster preservation, we use three query-oriented metrics.
First, the adjusted Rand index (ARI) measures global fidelity of the pairwise co-membership query \texttt{same\_halo($p_i$, $p_j$)} and adjusts for chance agreement. It is defined as

\begin{equation*}
    \mathrm{ARI}=\frac{2(TP\times TN - FP\times FN)}{(TP+FN)(FN+TN)+(TP+FP)(FP+TN)},
\end{equation*}
where $TP$, $TN$, $FP$, and $FN$ are computed over particle pairs according to halo co-membership. A true positive ($TP$) is a pair assigned to the same cluster in both the original and decompressed data, while a true negative ($TN$) is a pair assigned to different halos in both. A false positive ($FP$) occurs when two particles from different original clusters are merged into the same decompressed halo, and a false negative ($FN$) occurs when two particles from the same original cluster are split into different decompressed clusters. ARI equals $1$ for exact agreement, values near $0$ indicate chance-level agreement, and negative values indicate worse-than-chance agreement.

Second, intersection over union (IoU) measures preservation of the set-valued query \texttt{members(halo\_of($p_i$))}: given a particle, how accurately does the decompressed data return the member set of its original halo? Let $\{H_i\}$ and $\{H'_j\}$ denote the halos derived from the original and decompressed data, respectively. For each original halo $H_i$, we define
\begin{equation*}
    \mathrm{IoU}_i=\max_j
    \frac{|H_i\cap H'_j|}{|H_i\cup H'_j|}.
\end{equation*}
We summarize IoU using mass-weighted mean IoU (MWM-IoU), so larger halos contribute proportionally to their number of particles.

\begin{wrapfigure}{LH}{0.45\linewidth}
    \centering
    \includegraphics[width=\linewidth]{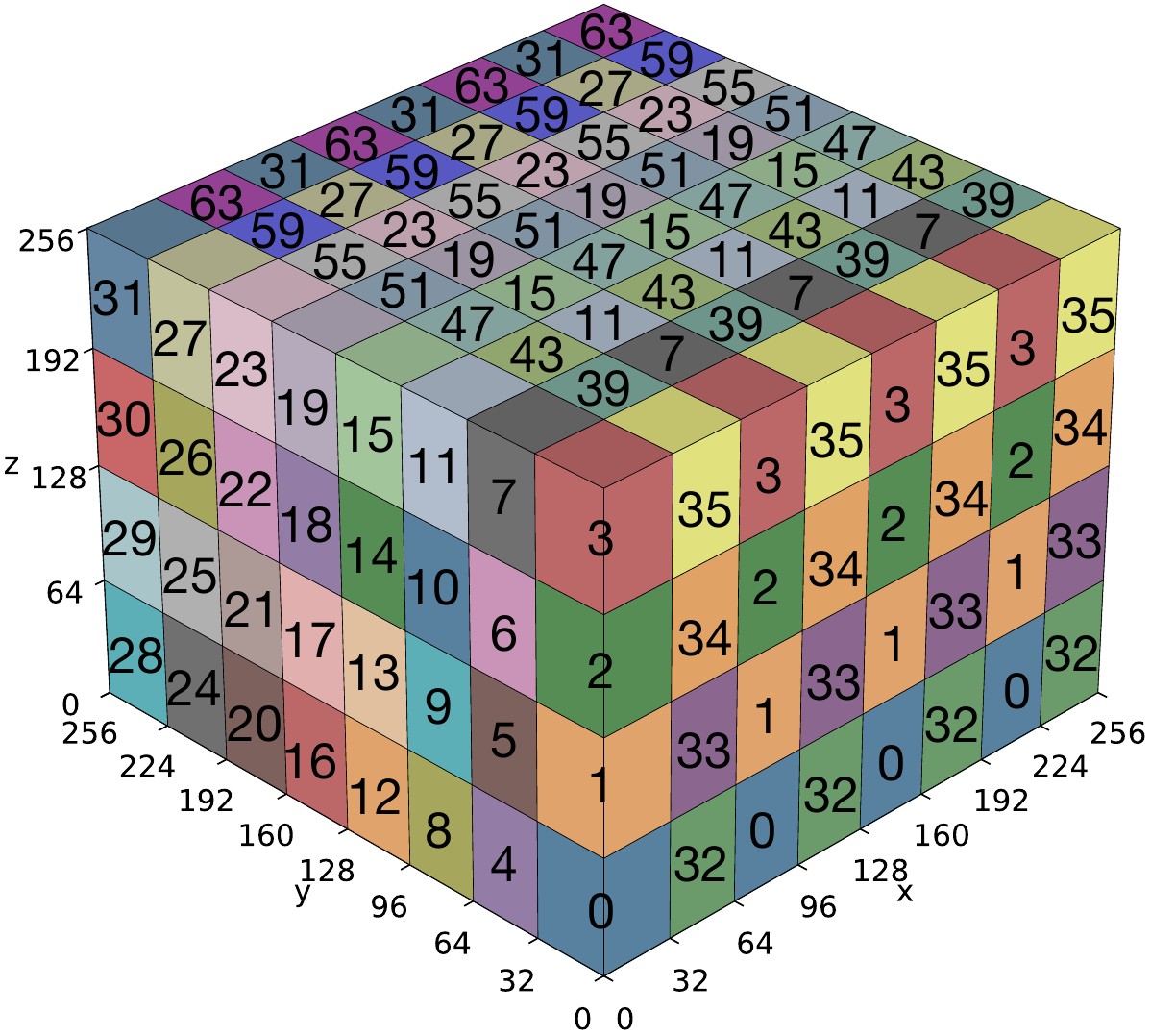}
    \caption{Spatial decomposition of the $256^3$ HACC simulation volume across 64 MPI ranks.
    Color-coded blocks represent individual rank assignments, with rank IDs annotated on visible faces.}
    \label{fig:hacc_distr}
\end{wrapfigure}

Finally, we report the halo mass function (HMF), an aggregate query over halos that counts halos by mass bin. HMF is a key downstream cosmology statistic for characterizing how many dark matter halos of a given mass exist within a volume of the universe.

\textbf{Datasets}. The datasets used in our evaluation, summarized in Table~\ref{tab:datasets}, span multiple scientific domains, spatial and temporal resolutions, and data modalities, and represents the diverse challenges faced by scientific data analysis. HACC data is generated from cosmological simulations carried out with the HACC framework~\cite{habib2013hacc,habib2016hacc}. EXAALT data comes from molecular dynamics simulations~\cite{exaalt}. Finite pointset method (FPM) datasets simulate the chemical process of salt dissolution in water, where higher-density salt diffuses into water to form unstable, downward-reaching structures known as viscous fingers~\cite{fpm}. For HACC (hiRes), results are reported on a single representative timestep; for FPM datasets, all available timesteps are evaluated and results are averaged, as the method's behavior is consistent across timesteps.

\begin{table}[!ht]
    \centering
    \footnotesize
    \caption{Benchmark Datasets. All datasets are in 3D, single-precision floating point.}
    \begin{tabular}{c|c|c|c}
    \toprule
        dataset & N (per timestep) & timesteps & size  \\ \midrule
        HACC (hiRes) & 1,073,734,015 & 16 & 192.00 GB \\ \hline
        HACC (lowRes) & 280,953,867 & 1 & 3.14 GB \\ \hline
        EXAALT & 2,869,440 & 1 & 32.84 MB \\ \hline
        FPM (hiRes) & 1,686,160 & 60 & 1.13 GB \\ \hline
        FPM (midRes) & 545,678 & 121 & 775.62 MB \\ \hline
        FPM (lowRes) & 196,066 & 121 & 271.50 MB \\
    \bottomrule
    \end{tabular}
    \label{tab:datasets}
\end{table}

For each HACC (hiRes) timestep, the $256^3$ domain is partitioned across 64 MPI ranks using a hierarchical, interleaved decomposition. While the $z$ and $y$ dimensions follow standard linear partitioning (4 and 8 segments, respectively), the $x$-dimension employs a non-contiguous strategy where each rank manages four disjoint sub-intervals. As shown in Fig.~\ref{fig:hacc_distr}, the domain is divided into 32 $yz$ ``super-blocks,'' each split along $x$ into two interleaved rank sets. This configuration creates spatially periodic ``slabs'' or ``pencils'' to facilitate long-range force calculations~\cite{habib2013hacc}.

\textbf{Hardware}. Experiments were conducted on the National Energy Research Scientific Computing Center (NERSC) Perlmutter supercomputer~\cite{perlmutter}. GPU-based evaluations used NVIDIA A100 (40 GB HBM2) nodes with CUDA 12.4, while CPU benchmarks ran on 64-core AMD EPYC 7763 nodes.

\subsection{Baseline comparison}

This section evaluates our proposed algorithm against state-of-the-art compressors, focusing on storage efficiency, throughput, and accuracy. Due to its substantial memory footprint exceeding a single GPU's capacity, all HACC (hiRes) results were obtained using a distributed configuration of 16 nodes (4 GPUs per node) on a single timestep unless otherwise specified. Peak GPU memory consumption per rank remains around 1.144 GB across all tested configurations, well within the 40 GB A100 capacity. Experiments for all other datasets were conducted on a single GPU. In PGD, we apply the adaptive moment estimation (Adam) algorithm~\cite{kingma2014adam} to adaptively adjust learning rate for faster convergence. Adam is selected for its fast convergence and memory efficiency, which are critical for processing large-scale data on GPU. Following standard practice, we use Adam with $\alpha=10^{-3}$, $\beta_1=0.9$, $\beta_2=0.999$, $\epsilon=10^{-8}$; results are robust to reasonable variation in $\alpha$ due to Adam's adaptive gradient scaling~\cite{kingma2014adam}. The linking threshold $b$ is specified through $\eta$; all experiments use the cosmological standard $\eta=0.2$.

\begin{figure}[!th]
    \centering
    \includegraphics[width=\linewidth]{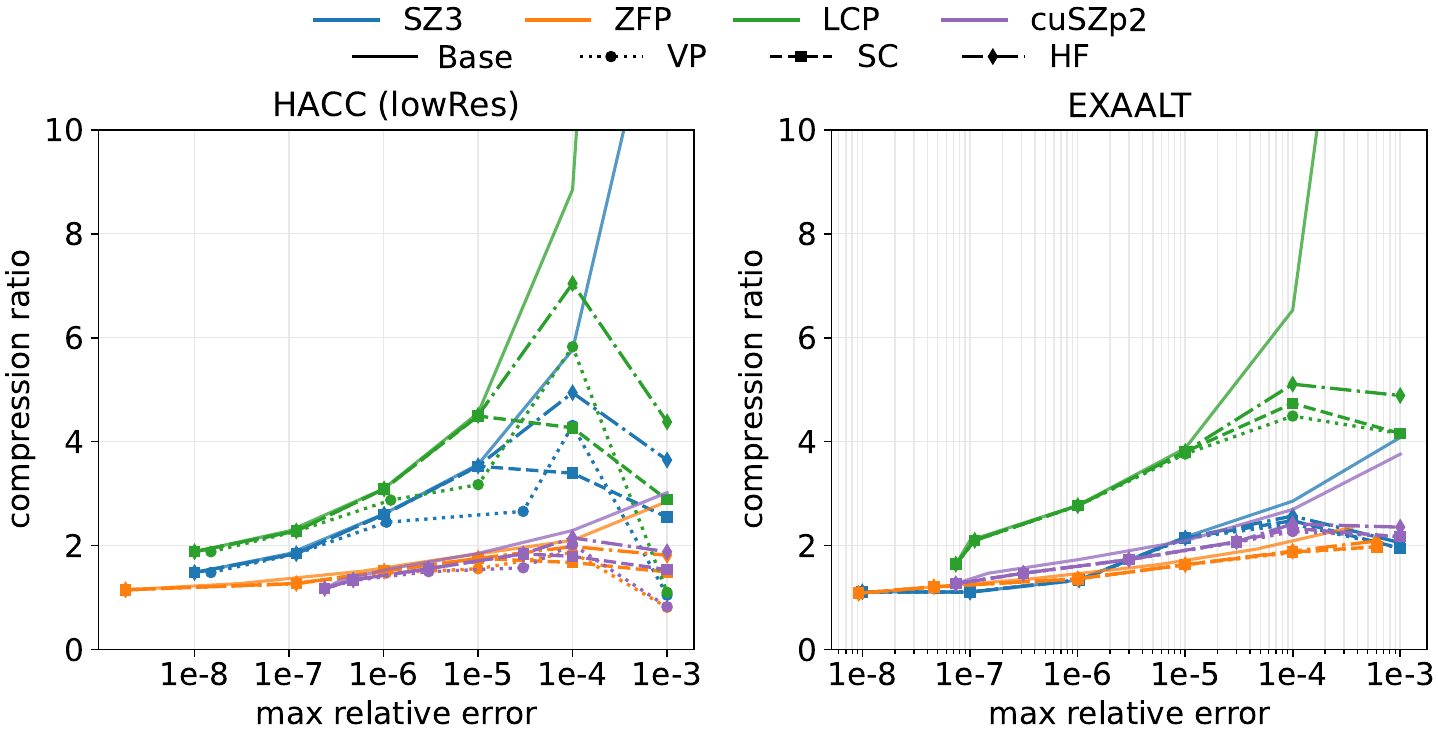}
    \caption{Compression ratio vs. maximum relative error for four base compressors and their VP-, SC-, and HF-corrected outputs on HACC (lowRes) and EXAALT.}
    \label{fig:cr_3modes}
\end{figure}

\begin{figure*}[!th]
    \centering
    \includegraphics[width=\linewidth]{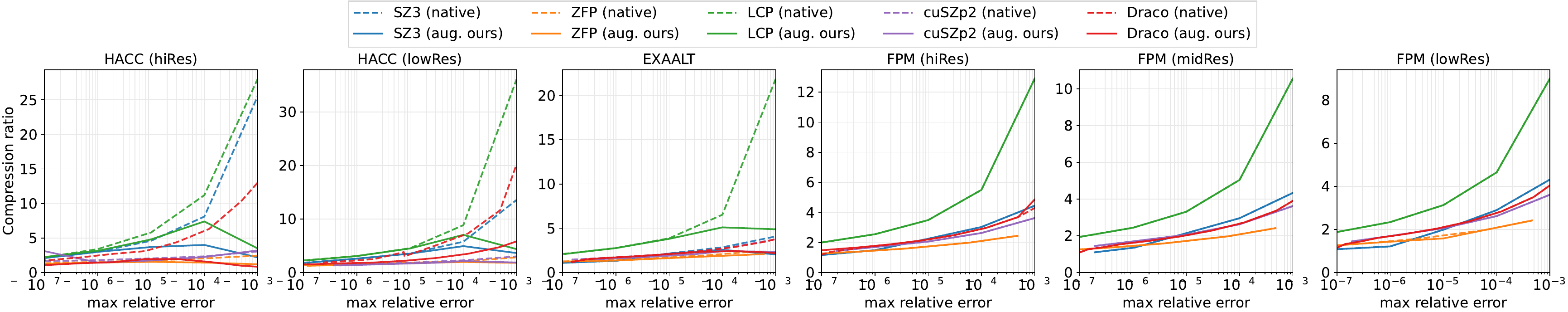}
    \caption{Compression ratio vs. maximum relative error of our method and baselines before and after correction on six datasets.}
    \label{fig:comp_ratio}
\end{figure*}

\begin{figure*}[!th]
    \centering
    \includegraphics[width=\linewidth]{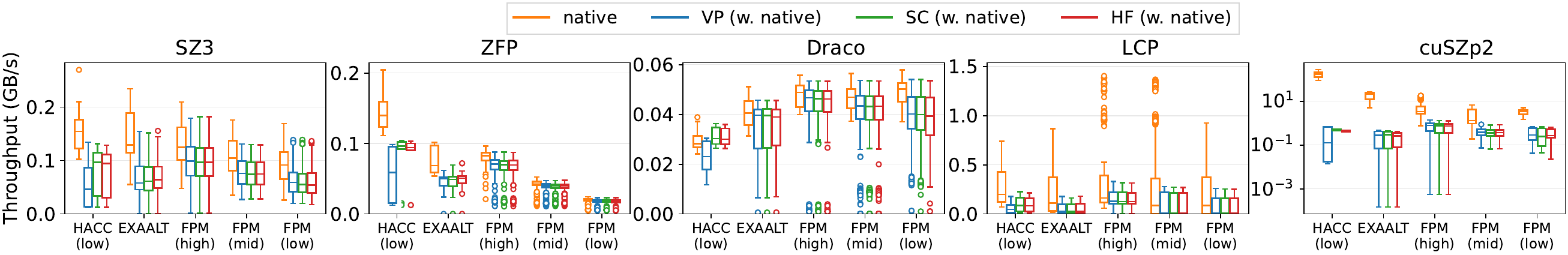}
    \caption{End-to-end throughput of base compressors with and without our correction.}
    \label{fig:throughput}
\end{figure*}

\textbf{Compression ratio}. We first compare all three constraint selection modes at full convergence under loss tolerance $\epsilon_\mathcal{L}=10^{-10}$, as shown in Fig.~\ref{fig:cr_3modes}. HF consistently achieves the highest compression ratio (i.e., smallest storage overhead), followed by SC and VP. This ordering follows from the size of the active-pair set $A$: a smaller set yields fewer editable particles and a shorter edit stream.

Fig.~\ref{fig:comp_ratio} shows the total compression ratio for each evaluated base compressor augmented with HF with full convergence under $\epsilon_\mathcal{L}=10^{-10}$. Among the candidates, LCP yields the highest compression ratio across all error bounds. The storage overhead required to preserve FoF clusters increases as the relative error bound loosens, whereas the base compressor's compression ratio improves with higher error tolerance. This competing dynamic creates an optimal point (or ``peak'') in the effective compression ratio, as seen in the HACC (hiRes), HACC (lowRes), and EXAALT datasets. For example, the total compression ratio for LCP on the HACC (lowRes) data peaks at a relative error of approximately $10^{-4}$. This suggests that if a baseline compression uses a coarser bound (e.g., $10^{-3}$), tightening the bound toward this peak value actually improves storage efficiency by reducing the subsequent correction overhead. Conversely, on FPM datasets, where vulnerable pairs are sparse, the storage overhead introduced by our correction is almost negligible, resulting in a monotonic increase in compression ratio without a discernible peak. The results on FPM datasets confirm that low violation density yields negligible overhead and demonstrate applicability beyond particle physics to fluid dynamics.

At relative errors of approximately $10^{-7}$ or smaller, the native and corrected curves overlap because no coordinate correction is needed: the base compressor already preserves cluster membership at these tight bounds. For LCP on the HACC datasets, the native compression ratio at the tight preservation bound is 1.88, whereas LCP augmented with HF reaches a cluster-preserving compression ratio around 7.5 at its peak, an approximately fourfold improvement.

\begin{wrapfigure}{LH}{0.58\linewidth}
    \centering
    \includegraphics[width=\linewidth]{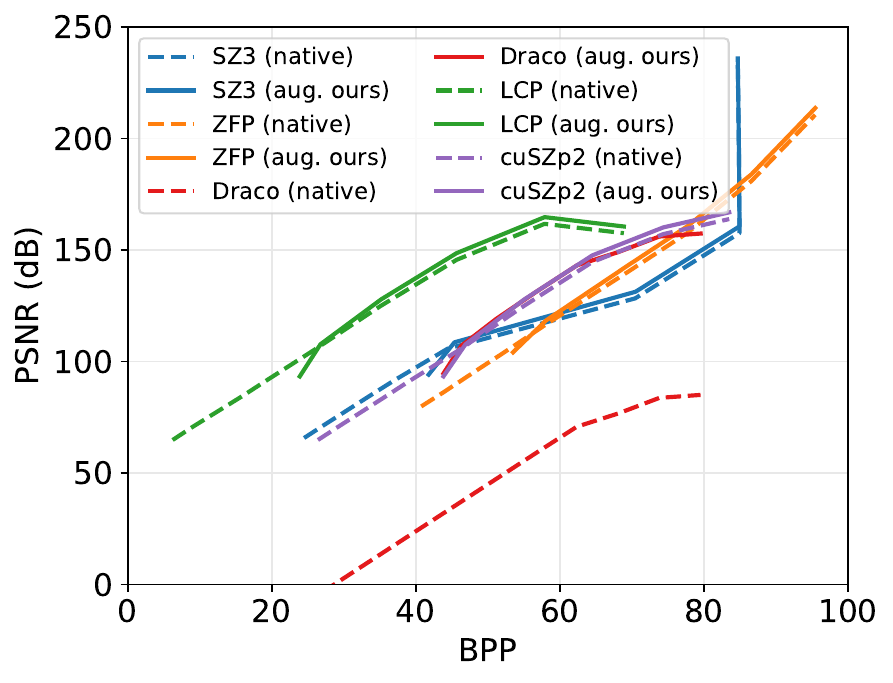}
    \caption{PSNR vs. BPP of our method and baselines before and after correction on EXAALT dataset.}
    \label{fig:rate-distortion}
\end{wrapfigure}

\textbf{Rate distortion}. We evaluate reconstruction accuracy versus storage cost by rate-distortion (RD) curves plotting PSNR against bits per particle (BPP) for HF mode in Fig.~\ref{fig:rate-distortion}. The reported BPP includes both the base compressed data and our edits, ensuring a fair comparison. Our correction shifts RD curves upward or leaves them unchanged, with a substantial gap (more than 50 dB) for Draco. Since our method optimizes for FoF cluster membership rather than pointwise accuracy and may perturb particles away from their base-compressed positions, the observed RD improvement is a beneficial byproduct that demonstrates our correction reclaims accuracy lost during initial compression without compromising cluster fidelity.

\textbf{Throughput}. The throughput of the base compressors and our method with full convergence is illustrated in Fig.~\ref{fig:throughput}. End-to-end throughput for our method includes base-compression time. Since HACC (hiRes) dataset needs a distributed MPI configuration while other datasets are processed on a single GPU, we provide a detailed scaling analysis for HACC (hiRes) in Section~\ref{sec:scaling} and focus here on single-node performance.

SC and HF add limited throughput overhead in most configurations. VP has particularly low throughput cases at loose error bounds because its larger active-pair and iteration counts make PGD dominant.
Table~\ref{tab:iteration} shows how active-pair count, iteration count, and union-find affect runtime on ZFP-compressed EXAALT data. Union-find adds little setup time, whereas reductions in active-pair and iteration counts substantially reduce PGD time, which dominates most configurations.
Furthermore, spatial partitioning and active-pair discovery impose an approximately 0.077\,s performance floor. This setup cost remains even with zero violations and iterations because the method must still discover active pairs and verify that no correction is required.

\begin{table}[!ht]
    \centering
    \footnotesize
    \caption{Convergence of the correction algorithm on the EXAALT dataset. Times in seconds; all violations are eliminated within reported iterations. The quantization step introduces no new violations because the safety margin $\epsilon_q$ in $\mathcal{L}_{\text{tight}}$ exceeds the maximum quantization error by construction. Double-precision is used for distances to accommodate small error bounds.}
    \begin{tabular}{c|c|c|c|c|c|c|c}
    \toprule
        rel. $\xi$ & mode & \# active prs & \# viol. prs & \# iter & $t_{total}$ & $t_{setup}$ & $t_{PGD}$ \\\midrule
        \multirow{3}{*}{$10^{-3}$} & VP & 41,445,603 & 305,855 & 114 & 0.877 & 0.124 & 0.753 \\ \cline{2-8}
         & SC & 41,445,603 & 305,855 & 114 & 0.856 & 0.131 & 0.725 \\ \cline{2-8}
         & HF & 36,120,322 & 210,250 & 83 & 0.647 & 0.152 & 0.495 \\ \hline
        \multirow{3}{*}{$10^{-4}$} & VP & 1,876,427 & 37,987 & 92 & 0.246 & 0.085 & 0.161 \\ \cline{2-8}
         & SC & 1,574,970 & 31,727 & 72 & 0.201 & 0.085 & 0.116 \\ \cline{2-8}
         & HF & 1,170,550 & 23,826 & 48 & 0.161 & 0.089 & 0.072 \\ \hline
        \multirow{3}{*}{$10^{-5}$} & VP & 185,048 & 4,785 & 5 & 0.079 & 0.075 & 0.0037 \\ \cline{2-8}
         & SC & 109,459 & 2,794 & 4 & 0.080 & 0.078 & 0.0020 \\ \cline{2-8}
         & HF & 105,039 & 2,658 & 4 & 0.105 & 0.103 & 0.0022 \\ \hline
        \multirow{3}{*}{$10^{-6}$} & VP & 18,508 & 313 & 2 & 0.073 & 0.072 & 0.0004 \\ \cline{2-8}
         & SC & 10,361 & 164 & 2 & 0.076 & 0.075 & 0.0003 \\ \cline{2-8}
         & HF & 10,321 & 162 & 2 & 0.098 & 0.097 & 0.0004 \\ \hline
        \multirow{3}{*}{$10^{-7}$} & VP & 1,808 & 34 & 1 & 0.070 & 0.070 & 0.0002 \\ \cline{2-8}
         & SC & 962 & 13 & 1 & 0.073 & 0.073 & 0.0002 \\ \cline{2-8}
         & HF & 962 & 13 & 1 & 0.077 & 0.077 & 0.0002 \\ \hline
        \multirow{3}{*}{$10^{-8}$} & VP & 200 & 0 & 0 & 0.074 & 0.074 & 0.0002 \\ \cline{2-8}
         & SC & 110 & 0 & 0 & 0.079 & 0.079 & 0.0002 \\ \cline{2-8}
         & HF & 110 & 0 & 0 & 0.077 & 0.077 & 0.0002 \\
    \bottomrule
    \end{tabular}
    \label{tab:iteration}
\end{table}

\begin{figure}
    \centering
    \includegraphics[width=\linewidth]{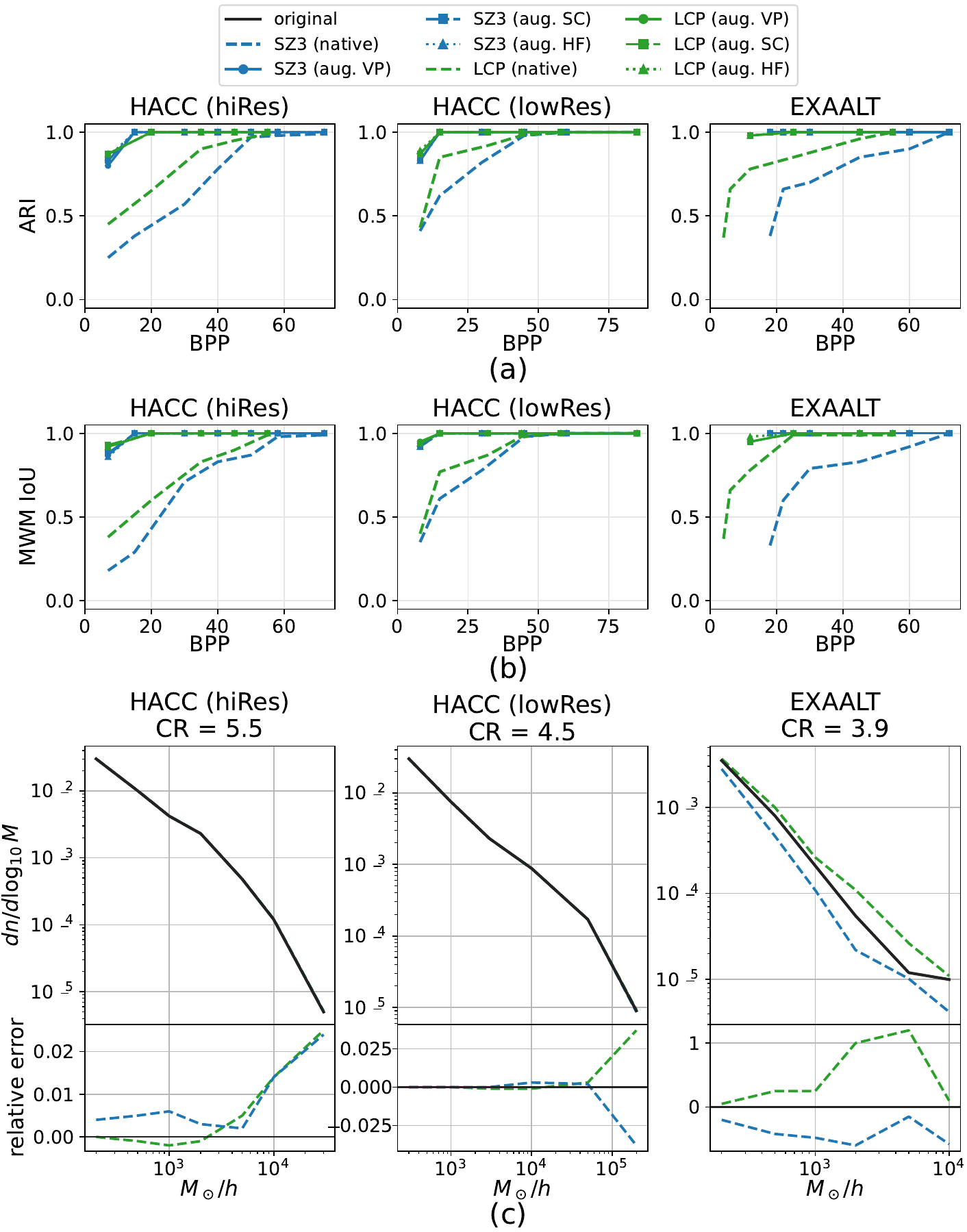}
    \caption{Accuracy of FoF clustering queries. (a) ARI of pairwise halo co-membership vs. BPP. (b) MWM-IoU of halo membership vs. BPP. (c) Halo mass functions (HMFs) and their pointwise relative errors for base compressors compared to our correction method at equivalent compression ratios. For $x$-values extending beyond the range of original data, relative error is calculated using linear interpolation of the original HMF.}
    \label{fig:fof-accuracy}
\end{figure}

\begin{wrapfigure}{LH}{0.33\linewidth}
    \centering
    \includegraphics[width=\linewidth]{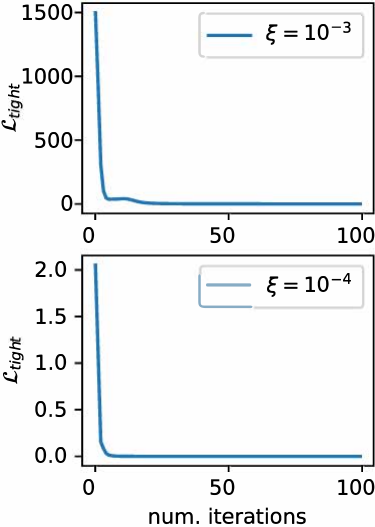}
    \caption{Tight loss $\mathcal{L}_{\text{tight}}$ vs. iteration count (limited to 100) for VP correcting ZFP-compressed EXAALT data with relative $\xi=10^{-3}$ (top) and $\xi=10^{-4}$ (bottom).}
    \label{fig:loss-vs-iter}
\end{wrapfigure}

\textbf{Adjusted Rand index (ARI) and intersection over union (IoU)}. Although our algorithm converges in all evaluations, the computational cost to reach full convergence may be prohibitive for datasets with high vulnerable-pair densities. For example, correcting ZFP-compressed HACC (lowRes) data with $\xi=10^{-3}$ requires 1,097 iterations to resolve 0.68 billion active pairs in HF mode, taking 35.62\,s, which is approximately 1.5 times of ZFP's compression time. However, as illustrated in Fig.~\ref{fig:loss-vs-iter}, the loss decreases sharply in the initial steps before entering a long, slow convergence tail in VP mode. Both factors, the high runtime cost of full convergence and the rapid early reduction in loss, suggest early termination as a viable heuristic. To evaluate the trade-off between execution time and clustering fidelity, we investigate the impact of a restricted iteration budget on FoF accuracy. We fix the maximum iteration count to 100 and plot ARI and mass-weighted mean (MWM) IoU against bit-rate in BPP in Fig.~\ref{fig:fof-accuracy}~{(a)}. Results for FPM datasets are omitted as the algorithm consistently converges within 100 iterations, yielding trivial ARI and MWM-IoU of 1.0.

Native base-compressor outputs exhibit low ARI and MWM-IoU on HACC (hiRes) because coordinate perturbations alter proximity links, including links that cross rank boundaries. Our distributed correction exchanges ghost particles to expose these boundary constraints. Although the 100-iteration cap can leave residual violations, the corrected outputs substantially improve both metrics, showing that truncated correction restores much of the FoF cluster structure.

\textbf{Halo mass function (HMF)}. To validate the efficacy of the truncated correction process for downstream FoF analysis, we calculate the HMFs for results using a 100-iteration limit; as Table~\ref{tab:iteration} shows, most configurations converge in far fewer iterations, and the cap is only approached under the most aggressive error bounds where residual loss is already negligible. Fig.~\ref{fig:fof-accuracy}~{(b)} compares the HMFs ($B=50$ bins) derived from SZ3 and LCP, the base compressors with the highest compression ratios, against our correction results. The impact of our algorithm is most evident in the lower panel, which displays the pointwise relative error of the reconstructed HMFs. At the same compression ratios, our method consistently reduces the relative error across the whole HMFs, more accurately preserving the statistical distribution of the cosmological halos.

\begin{table*}[!th]
    \centering
    \scriptsize
    \caption{Single-node CUDA kernels and CPU functions performance metrics on EXAALT dataset with LCP at $\xi=10^{-4}$.}
    \label{tab:single_node_perf}
    \resizebox{\textwidth}{!}{
    \begin{tabular}{c|c|c|c|c|c|c|c|c}
        \toprule
        \textbf{Kernel / Function} & \textbf{Mode} & \textbf{Calls} & \textbf{CPU Time} & \textbf{GPU Time} & \textbf{GPU BW (GB/s)} & \textbf{GPU BW Eff. (\%)} & \textbf{GPU AI (F/B)} & \textbf{Speedup} \\
        \midrule
        \multirow{3}{*}{\texttt{spatialPartitioning}} & VP & 1 & 1.754\,s & 824.3\,$\mu$s & 519.2 & 33.4 & 0.225 & 2128$\times$ \\ \cline{2-9}
         & SC & 1 & 1.773\,s & 830.6\,$\mu$s & 511.8 & 32.9 & 0.226 & 2134$\times$ \\ \cline{2-9}
         & HF & 1 & 1.772\,s & 824.7\,$\mu$s & 530.7 & 34.1 & 0.225 & 2149$\times$ \\
        \hline
        \multirow{3}{*}{\texttt{findVulnerablePairs}} & VP & 1 & 10.535\,s & 76.45\,ms & 39.0 & 2.5 & 0.100 & 137.8$\times$ \\ \cline{2-9}
         & SC & 1 & 10.539\,s & 79.15\,ms & 49.2 & 3.2 & 0.095 & 133.1$\times$ \\ \cline{2-9}
         & HF & 1 & 10.565\,s & 80.16\,ms & 52.0 & 3.3 & 0.088 & 131.8$\times$ \\
        \hline
        \multirow{3}{*}{\texttt{filterConstraints}} & VP & 0 & - & - & - & - & - & - \\ \cline{2-9}
         & SC & 1 & 11.929\,s & 527.1\,$\mu$s & 340.8 & 21.9 & 0.019 & 22630$\times$ \\ \cline{2-9}
         & HF & 1 & 12.330\,s & 624.9\,$\mu$s & 341.7 & 22.0 & 0.000 & 19731$\times$ \\
        \hline
        \multirow{3}{*}{\texttt{PGDComputeLoss}} & VP & 100 & 877.38\,ms & 16.79\,ms & 934.3 & 60.1 & 0.387 & 52.3$\times$ \\ \cline{2-9}
         & SC & 90 & 600.20\,ms & 13.90\,ms & 956.0 & 61.5 & 0.355 & 43.2$\times$ \\ \cline{2-9}
         & HF & 65 & 343.12\,ms & 8.98\,ms & 958.4 & 61.6 & 0.300 & 38.2$\times$ \\
        \hline
        \multirow{3}{*}{\texttt{PGDComputeGradients}} & VP & 99 & 876.14\,ms & 16.76\,ms & 967.1 & 62.2 & 0.306 & 52.3$\times$ \\ \cline{2-9}
         & SC & 89 & 595.96\,ms & 13.80\,ms & 980.7 & 63.1 & 0.282 & 43.2$\times$ \\ \cline{2-9}
         & HF & 64 & 338.92\,ms & 8.87\,ms & 1013 & 65.1 & 0.238 & 38.2$\times$ \\
        \hline
        \multirow{3}{*}{\texttt{PGDUpdatePositions}} & VP & 99 & 6.864\,s & 131.32\,ms & 810.1 & 52.1 & 0.124 & 52.3$\times$ \\ \cline{2-9}
         & SC & 89 & 4.821\,s & 111.62\,ms & 811.4 & 52.2 & 0.123 & 43.2$\times$ \\ \cline{2-9}
         & HF & 64 & 2.876\,s & 75.31\,ms & 810.6 & 52.1 & 0.000 & 38.2$\times$ \\
        \hline
        \multirow{3}{*}{\texttt{losslesslyCompressEdits}} & VP & 6 & 3.452\,s & 459.8\,$\mu$s & 6.5 & 0.4 & 0.000 & 7509$\times$ \\ \cline{2-9}
         & SC & 6 & 3.318\,s & 438.0\,$\mu$s & 12.6 & 0.8 & 0.000 & 7575$\times$ \\ \cline{2-9}
         & HF & 6 & 3.224\,s & 411.2\,$\mu$s & -- & -- & -- & 7840$\times$ \\
        \hline
        \multirow{3}{*}{\textbf{end-to-end}} & VP & 1 & 25.640\,s & 454.92\,ms & -- & -- & -- & 56.4$\times$ \\ \cline{2-9}
         & SC & 1 & 33.576\,s & 400.55\,ms & -- & -- & -- & 83.8$\times$ \\ \cline{2-9}
         & HF & 1 & 31.449\,s & 338.33\,ms & -- & -- & -- & 93.0$\times$ \\
        \bottomrule
    \end{tabular}}
\end{table*}

\subsection{Single-node performance}
Table~\ref{tab:single_node_perf} details per-kernel performance for correction process on LCP-compressed EXAALT data ($\xi=10^{-4}$), involving 1.8M vulnerable pairs. The GPU achieves $56\times-93\times$ end-to-end speedup over the 64-core CPU baseline, depending on the constraint-selection mode. HF mode achieves the highest speedup because its smaller active-pair set reduces both the number of iterations and the work per iteration.

All computational phases exhibit arithmetic intensity (AI) significantly below the A100 ridge points (9.75 F/B), confirming that the algorithm is memory-bound and that performance is governed by sustained bandwidth rather than compute throughput. The three PGD kernels, which dominate iteration time, reach 52-65\% of the A100's peak HBM2 bandwidth (2 TB/s). The \texttt{PGDUpdatePositions} kernel assigns one thread per editable particle, enabling embarrassingly parallel Adam updates and box-constrained projections. The \texttt{PGDComputeLoss} kernel uses shared-memory binary-tree reduction to accumulate pair violations without global atomic traffic, sustaining near-peak bandwidth.

\begin{table*}[t]
\centering
\scriptsize
\caption{Performance breakdown and load imbalance across scales for CPU and GPU MPI implementations on HACC (hiRes), HF mode. Times are maximum seconds across ranks; imbalance is $(t_{max}-t_{avg})/t_{avg}$.}
\label{tab:multi_node_perf}
\resizebox{\textwidth}{!}{
\begin{tabular}{l|l|c|c|c|c|c|c|c|c|c|c|c|c}
\toprule
 & & \multicolumn{3}{c|}{\textbf{CPU Time (sec)}} & \multicolumn{3}{c|}{\textbf{CPU Imbalance}} & \multicolumn{3}{c|}{\textbf{GPU Time (sec)}} & \multicolumn{3}{c}{\textbf{GPU Imbalance}} \\
\cmidrule(lr){3-5} \cmidrule(lr){6-8} \cmidrule(lr){9-11} \cmidrule(lr){12-14}
\textbf{Group} & \textbf{Phase} & \textbf{16p} & \textbf{64p} & \textbf{256p} & \textbf{16p} & \textbf{64p} & \textbf{256p} & \textbf{16p} & \textbf{64p} & \textbf{256p} & \textbf{16p} & \textbf{64p} & \textbf{256p} \\
\midrule
\multirow{3}{*}{\textbf{I/O}}
 & Read original & 1.590 & 0.2198 & 0.1892 & 7.8\% & 16.8\% & 43.0\% & 0.0743 & 0.0219 & 0.0076 & 9.7\% & 20.9\% & 42.9\% \\
 & Read decompressed & 1.513 & 0.2376 & 0.1624 & 6.5\% & 33.4\% & 38.2\% & 0.5844 & 0.1342 & 0.0458 & 6.3\% & 16.0\% & 47.5\% \\
 & Write edits & 0.0032 & 0.0430 & 0.0128 & 28.8\% & 1429.5\% & 947.4\% & 0.0047 & 0.0029 & 0.0020 & 24.0\% & 35.0\% & 266.5\% \\
\midrule
\multirow{3}{*}{\textbf{Comm.}}
 & BBox exchange & 0.3337 & 0.2272 & 0.6937 & 101.3\% & 118.0\% & 65.1\% & 0.0828 & 0.0816 & 0.0859 & 297.7\% & 102.6\% & 55.1\% \\
 & Ghost exchange & 7.612 & 4.088 & 0.7497 & 2.6\% & 12.0\% & 37.1\% & 2.479 & 0.8293 & 0.2045 & 3.9\% & 10.2\% & 10.8\% \\
 & Convergence check & 93.40 & 57.22 & 30.52 & 98.2\% & 88.4\% & 53.2\% & 0.0836 & 0.0381 & 0.0212 & 96.0\% & 88.5\% & 50.6\% \\
\midrule
\multirow{7}{*}{\makecell[l]{\textbf{Local}\\\textbf{Compute}}}
 & BBox compute & 3.703 & 1.811 & 0.3627 & 15.3\% & 23.7\% & 62.7\% & 0.0322 & 0.0532 & 0.0303 & 39.4\% & 331.3\% & 519.1\% \\
 & Grid partition & 122.9 & 36.78 & 5.412 & 16.3\% & 15.1\% & 25.3\% & 0.1474 & 0.1174 & 0.0653 & 61.6\% & 78.0\% & 323.3\% \\
 & VP detection & 504.7 & 141.8 & 36.76 & 5.3\% & 9.7\% & 54.9\% & 0.9031 & 0.3522 & 0.1392 & 12.0\% & 76.0\% & 249.5\% \\
 & Union-find & 555.6 & 156.6 & 37.05 & 4.5\% & 8.8\% & 49.6\% & 0.6237 & 0.1265 & 0.0343 & 8.1\% & 27.9\% & 140.6\% \\
 & PGD optimization & 10.02 & 3.243 & 0.9426 & 11.6\% & 24.1\% & 56.4\% & 0.7424 & 0.2921 & 0.0454 & 23.6\% & 39.9\% & 47.4\% \\
 & Edit encoding & 255.5 & 69.26 & 25.44 & 4.8\% & 13.3\% & 49.1\% & 0.0925 & 0.0666 & 0.0438 & 54.0\% & 199.5\% & 260.5\% \\
 & Cleanup & 0.0127 & 0.0045 & 0.0021 & 11.8\% & 16.6\% & 33.4\% & 0.0127 & 0.0484 & 0.0325 & 164.0\% & 180.8\% & 1212.2\% \\
\midrule
\multicolumn{2}{l|}{\textbf{H2D / D2H}} & --- & --- & --- & --- & --- & --- & 0.2803 & 0.0959 & 0.0426 & 5.2\% & 29.8\% & 148.4\% \\
\midrule
\multicolumn{2}{l|}{\textbf{Total Wall Clock}} & \textbf{1496.3} & \textbf{427.5} & \textbf{114.04} & \textbf{1.2\%} & \textbf{2.0\%} & \textbf{4.3\%} & \textbf{6.103} & \textbf{1.888} & \textbf{0.6281} & \textbf{1.2\%} & \textbf{3.5\%} & \textbf{8.8\%} \\
\midrule
\multicolumn{2}{l|}{\textbf{End-to-end GPU Speedup}} & \multicolumn{6}{c|}{---} & \textbf{245.2$\times$} & \textbf{226.4$\times$} & \textbf{181.6$\times$} & \multicolumn{3}{c}{---} \\
\bottomrule
\end{tabular}}
\end{table*}

\subsection{Inter-node distributed scaling}
\label{sec:scaling}

\begin{figure}[!ht]
    \centering
    \includegraphics[width=\linewidth]{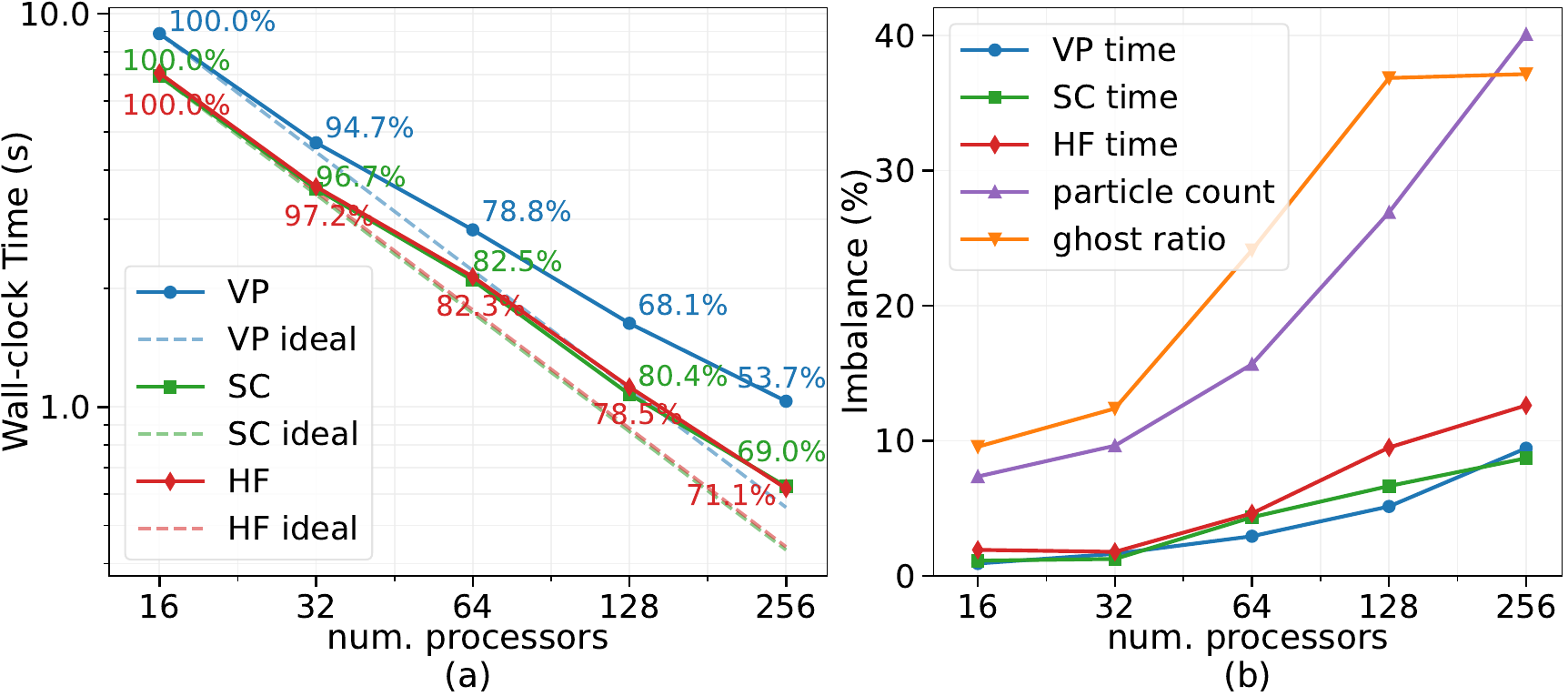}
    \caption{Strong scaling analysis for the GPU implementation on the HACC (hiRes) dataset. (a) Wall-clock time (blue line) and parallel efficiency (texts) relative to the 16-process baseline; the dashed line represents ideal linear scaling. (b) Analysis of system imbalance across three metrics: execution time per rank, local particle distribution, and the ghost-to-local particle ratio.}
    \label{fig:strong_scaling}
\end{figure}

We evaluate the CPU-MPI and GPU-MPI implementations to measure distributed-memory coordination overhead and scalability.

\textbf{Strong scaling}. Fig.~\ref{fig:strong_scaling} and Table~\ref{tab:multi_node_perf} present a strong scaling analysis of our correction algorithm on SZ3-compressed HACC (hiRes) dataset with relative $\xi=10^{-6}$ in overall and breakdown senses, respectively. To evaluate 16-256 processes while maintaining HACC's geometric logic as in in-situ compression and correction, we aggregate data by merging ranks with adjacent indices into larger units (e.g., grouping ranks 0–3 into a single process for a 16-process run). Conversely, for 128- and 256-process configurations requiring higher granularity, the data within each rank is subdivided along the $x$-axis. Since HACC's spatial decomposition changes with rank count (adjacent ranks merged for fewer nodes, sub-divided for more), scaling results reflect both parallel efficiency and decomposition effects; a fixed-decomposition strong scaling test would require re-partitioning the simulation data, which is outside the scope of this work. Correctness is preserved across all scales: the ghost zone width $\delta = b + 2\sqrt{3}\xi$ guarantees every boundary-crossing halo and its PGD gradient are fully replicated, making the results invariant to process count.

Table~\ref{tab:multi_node_perf} shows that the GPU implementation achieves end-to-end speedups of 245.2$\times$, 226.4$\times$, and 181.6$\times$ over the CPU at 16, 64, and 256 ranks, respectively. Scaling also differs by constraint-selection mode: VP efficiency falls to 53.7\% at 256 ranks, whereas SC and HF retain approximately 70\% (Fig.~\ref{fig:strong_scaling}). Their smaller active-pair sets therefore improve both runtime and parallel efficiency.

The two implementations scale differently. The CPU version retains 82\% efficiency at 256 ranks because its runtime is dominated by well-scaling local computation. By contrast, the GPU version reduces local computation to milliseconds, making communication a larger fraction of runtime, especially at 256 ranks.
From 16 to 256 ranks, local compute remains stable at roughly 50\% of runtime, while communication grows from 42.2\% to 49.0\%.

Imbalance analysis in Fig.~\ref{fig:strong_scaling} and Table~\ref{tab:multi_node_perf} shows that total wall-clock imbalance remains modest at all scales (at most 8.8\% on GPU and 4.3\% on CPU), even when individual phases are highly skewed. The convergence check shows 50--98\% imbalance, largely due to waiting for the slowest rank, while short phases such as bounding-box computation, edit writing, and cleanup can show percentage imbalances in the hundreds but contribute negligible absolute time. Fig.~\ref{fig:strong_scaling}~(b) associates the growing skew with particle-count and ghost-ratio imbalance, which reach roughly 40\% at 256 ranks.

\textbf{Weak scaling}. To simulate exascale-class workloads, we concatenate HACC (hiRes) timesteps along the $x$-axis, maintaining a constant per-rank workload. The problem scales from 16 nodes (64 ranks, 1 dataset) up to 256 nodes (1,024 ranks, 16 datasets), increasing the simulation domain from $256^3$ to $4,096 \times 256 \times 256$. While not physically equivalent to larger simulation volumes, this data construction of increasing size isolates how correction cost scales with $N$, independent of variations in clustering structure or simulation parameters.

\begin{figure}[!ht]
    \centering
    \includegraphics[width=\linewidth]{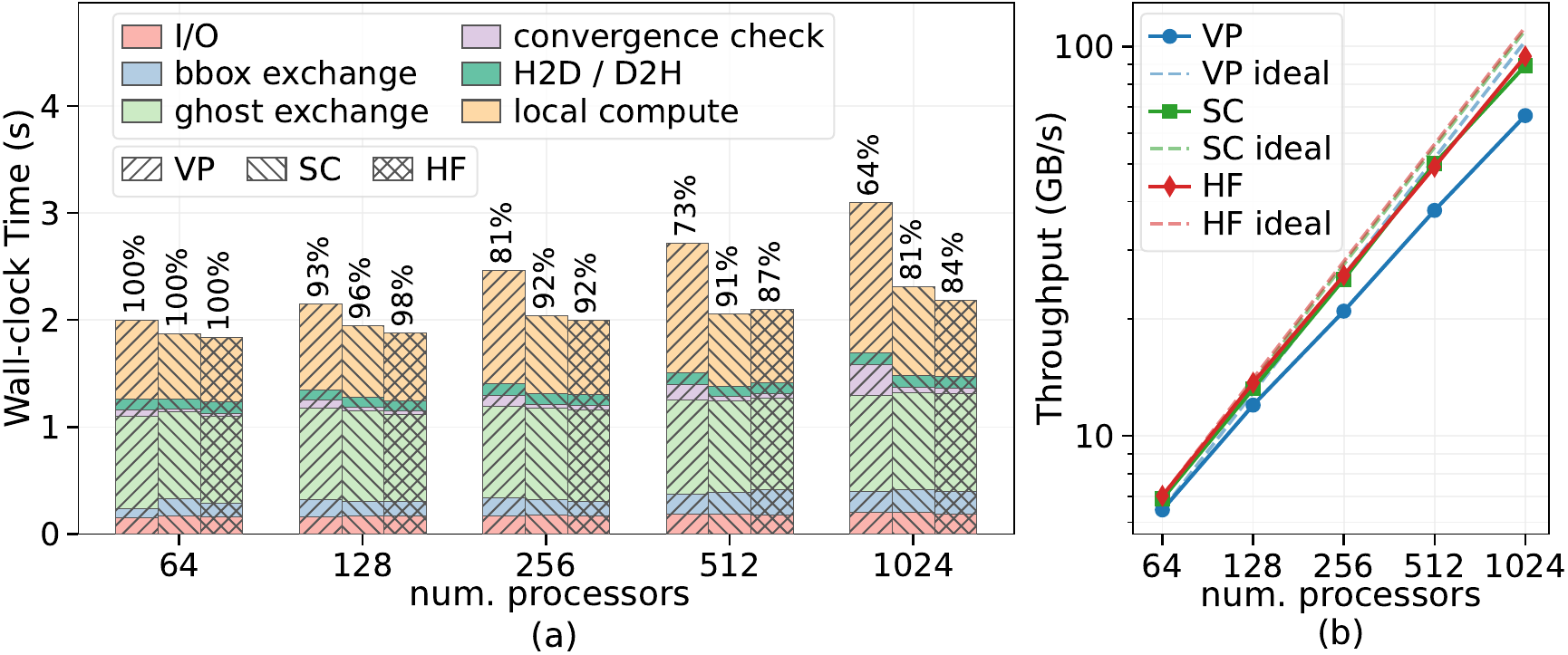}
    \caption{Weak scaling execution time breakdown and throughput for the GPU implementation. (a) Wall-clock time breakdown across scales. Each stacked bar shows the maximum time over all ranks. Percentages on the bars are overall weak scaling efficiency. (b) The aggregate throughput. The dashed line represents ideal linear scaling.}
    \label{fig:weak_scaling}
\end{figure}

Fig.~\ref{fig:weak_scaling} shows the execution time breakdown, weak scaling efficiency, and aggregate throughput as the problem scales from 64 to 1,024 ranks. At 1,024 ranks, HF and SC sustain 84\% and 81\% weak-scaling efficiency, respectively, while VP reaches 64\% (Fig.~\ref{fig:weak_scaling}~(a)); aggregate throughput closely follows the ideal linear-scaling curve across the tested range (Fig.~\ref{fig:weak_scaling}~(b)). The breakdown in Fig.~\ref{fig:weak_scaling}~(a) shows that local computation per rank remains nearly constant, while additional wall-clock time is concentrated in communication and synchronization as the rank count grows.

Table~\ref{tab:weak_scaling} shows that data skew and time skew decouple at scale. Data imbalance nearly doubles, from 12.22\% at 64 ranks to 24.69\% at 1,024 ranks, yet time imbalance remains between roughly 3\% and 7\% without monotonic growth (e.g., 3.97\% for HF and 4.16\% for VP at 1,024 ranks). Thus, for the tested decompositions, greater particle-count imbalance does not produce comparable execution-time imbalance.

\begin{table}[!ht]
    \centering
    \footnotesize
    \caption{Performance and workload for weak scaling.}
    \begin{tabular}{c|c|c|c|c|c}
        \toprule
        Proc. & \makecell{Avg. Particles\\per Rank} & \makecell{Data\\Imbalance} & \makecell{VP Time\\Imbalance} & \makecell{SC Time\\Imbalance} & \makecell{HF Time\\Imbalance} \\ \midrule
        64 & 16,777,117 & 12.22\% & 3.12\% & 6.23\% & 5.20\% \\ \hline
        128 & 16,777,115 & 12.93\% & 4.48\% & 5.72\% & 3.77\% \\ \hline
        256 & 16,777,109 & 14.54\% & 4.56\% & 6.14\% & 5.36\% \\ \hline
        512 & 16,777,101 & 16.56\% & 3.28\% & 6.93\% & 6.24\% \\ \hline
        1,024 & 16,777,083 & 24.69\% & 4.16\% & 5.70\% & 3.97\% \\
        \bottomrule
    \end{tabular}
    \label{tab:weak_scaling}
\end{table}

%% file: 5.conclusion.tex
\section{Conclusion}

We presented a correction method that preserves FoF cluster membership queries under error-bounded lossy compression. The method formulates cluster preservation as constrained coordinate optimization and uses three constraint-selection modes that exploit connected-component semantics to reduce the enforced pair set. Its GPU and distributed-memory implementations work with off-the-shelf base compressors. Across cosmology, molecular dynamics, and fluid dynamics datasets, all uncapped runs recover exact cluster membership; with a 100-iteration cap, correction substantially improves ARI, MWM-IoU, and HMF accuracy. The best configuration achieves an approximately fourfold higher compression ratio than the same base compressor configured with the tight error bound needed to preserve clustering, while maintaining competitive end-to-end throughput.

\textbf{Limitations}. Each correction stream in our evaluation targets one linking length $b$, so a query at a different $b$ requires a new correction of the base-decompressed data. A range of thresholds can be supported by enforcing the union of their selected active pairs, but the additional constraints and metadata reduce compression ratio and throughput. At loose relative error bounds ($\xi \geq 10^{-3}$), correction can exceed base-compression time when vulnerable-pair density is high. The current implementation also assumes a non-periodic domain and does not support 6D phase-space finders such as Rockstar.

\textbf{Future works}. We plan to add periodic-boundary support, exploit temporal coherence in time-evolving datasets to reduce the edit footprint, generalize the loss to phase-space halo finders such as Rockstar, and integrate correction into the base compressor's transform stage to eliminate the separate edit stream.